\documentclass[10pt,twocolumn,letterpaper]{article}

\usepackage[pagenumbers]{cvpr}
\usepackage{multirow}
\usepackage{placeins}
\usepackage{makecell}
\usepackage{bbding}
\usepackage{enumitem}
\usepackage{colortbl}
\usepackage{xcolor}
\usepackage{balance}
\usepackage{listings}
\usepackage{tikz}
\usetikzlibrary{arrows.meta,positioning}
\definecolor{myblue}{RGB}{233, 241, 249}
\definecolor{mygray}{RGB}{99, 110, 114}
\definecolor{myred}{RGB}{255, 118, 117}
\definecolor{myyellow}{RGB}{255, 234, 167}
\definecolor{mygreen}{RGB}{216, 226, 204}

\definecolor{cvprblue}{rgb}{0.21,0.49,0.74}
\usepackage[pagebackref,breaklinks,colorlinks,allcolors=cvprblue]{hyperref}
\hypersetup{
    pdftitle={SCAP: Stochastic Context for Action Expert Pretraining},
    pdfauthor={Xiao Lin, Yaoxu Lyu, Ronghao Dang, Zhili Liu, Kaijing Ma, Yang Zhou, Dingning Liu, Runzhe Teng, Jun Cen, Shanghang Zhang, Chengju Liu, Qijun Chen},
    pdfsubject={Preprint}
}

\def\confName{CVPR}
\def\confYear{2027}

\usepackage{CJKutf8}

\title{SCAP: Stochastic Context for Action Expert Pretraining}
\author{
Xiao Lin$^{1,*}$ \hspace{0.9em} Yaoxu Lyu$^{2,4,*}$ \hspace{0.9em} Ronghao Dang$^{4,*,\dagger}$ \hspace{0.9em} Zhili Liu$^{4}$ \hspace{0.9em} Kaijing Ma$^{3,4}$ \hspace{0.9em} Yang Zhou$^{3,4}$\\
Dingning Liu$^{3,4}$ \hspace{0.9em} Runzhe Teng$^{3,4}$ \hspace{0.9em} Jun Cen$^{4}$ \hspace{0.9em} Shanghang Zhang$^{2}$ \hspace{0.9em} Chengju Liu$^{1}$ \hspace{0.9em} Qijun Chen$^{1}$\\[0.5ex]
$^{1}$Tongji University, $^{2}$Peking University, $^{3}$Fudan University, $^{4}$JD Future Academy
}

\begin{document}
\maketitle
\begingroup
\renewcommand{\thefootnote}{\fnsymbol{footnote}}
\footnotetext[1]{Equal contribution.}
\footnotetext[2]{Corresponding author: Ronghao Dang.}
\endgroup
\begin{abstract}
Single-modality action-expert pretraining controls competition from visual shortcuts before vision--language--action (VLA) adaptation.
Yet single-modality pretraining can impair the vision--language interactions needed for subsequent joint prediction.
We construct a diagnostic model in which VA or LA pretraining can disable an initially active pathway for joint vision--language modeling.
In this model, subsequent training with complete visual and language inputs cannot restore the pathway, even with unlimited adaptation.
We therefore introduce \textbf{Stochastic Context for Action Expert Pretraining (SCAP)}, a new action-expert pretraining stage that precedes full-input training.
During this stage, visual and text tokens are randomly masked before entering the VLM.
In this unified diagnostic model, SCAP can preserve the joint vision--language modeling pathway during pretraining and subsequent full-input training.
Compared with direct full-input training, it can also reduce reliance on visual shortcuts and increase reliance on language instructions.
SCAP consistently improves performance across four different VLA architectures.
In real-robot experiments, it achieves 76.67\% average success across three instruction-conditioned manipulation tasks, outperforming full-input pretraining by 25 percentage points.
\end{abstract}

\section{Introduction}
\label{sec:intro}

\begin{figure}[t]
    \centering
    \includegraphics[width=\columnwidth]{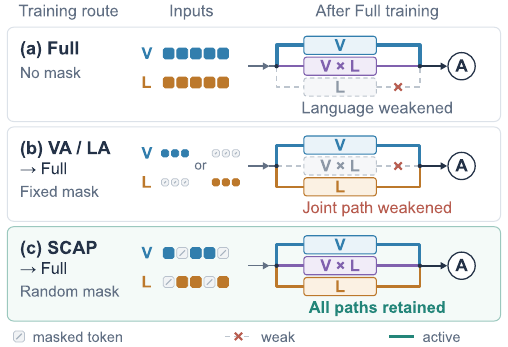}
    \caption{\textbf{Pretraining shapes multimodal modeling pathways.}}
    \label{fig:scap_motivation}
\end{figure}

% 逻辑：从 VLA 的目标切入，用两篇诊断工作说明视觉 shortcut 问题，并把动作专家预训练定位为干预点。
%v1 % Vision--language--action (VLA) models build robot policies on pretrained vision--language models (VLMs), using an action expert to turn the observed scene and the instruction into continuous actions~\cite{black2024pi0}. Following instructions requires actions to depend on language, yet demonstrations rarely enforce this: when the scene alone predicts the demonstrated behavior, a policy can fit the data through visual cues and ignore the instruction. Recent studies trace such visual shortcuts to demonstrations in which scenes strongly predict instructions~\cite{lian2026langforce}, and show that fine-tuning can make policies blind to instructions~\cite{zhang2026flatnesspreservesinstructionfollowing}. Because the action expert acquires its input dependencies during training, how it is pretrained is a natural place to intervene.

Vision--language--action (VLA) models build robot policies on pretrained vision--language models (VLMs), using an action expert to turn the observed scene and the instruction into continuous actions~\cite{black2024pi0}. Recent embodied foundation models further strengthen spatial grounding and planning through specialized pretraining and adaptation~\cite{dang2026rynnbrain,li2026rynnbrain11,zhu2026rynnunify}. However, scene--instruction correlations in demonstrations can encourage visual shortcuts, allowing policies to ignore language~\cite{lian2026langforce,zhang2026flatnesspreservesinstructionfollowing}. Because the action expert is randomly initialized, it can readily inherit these biases and may, in turn, degrade the pretrained VLM's representations through backpropagation during joint fine-tuning. Action expert pretraining is thus worth exploring to foster language dependence.

% 逻辑：先 LA 及代表作，再 VA 及代表作；承认两者是完整系统，只抽出本文研究的共同设计选择。
Existing recipes counter visual shortcuts by controlling which modality the expert sees. APT~\cite{xu2026apt} and BayesVLA~\cite{xu2025bayesvla} adopt vision--action (VA) pretraining, first learning a visuomotor prior and then introducing language. Conversely, LA4VLA~\cite{lin2026la4vla} uses language--action (LA) pretraining without vision to learn instruction--action associations, exploring both sequential and mixed LA--VLA training. Despite their differences, these methods follow the same paradigm of incorporating a unimodal pretraining stage preceding full pretraining.

% 逻辑：用真机配杯任务直观说明联合依赖，再给出反例结论：单模态阶段可关闭联合通路，后续完整输入训练无法恢复。
%v1 % However, many instructions can only be grounded by combining the two modalities. In ``place a cup matching the plate at the seat near the teapot,'' the instruction does not reveal which seat is near the teapot or what color its plate is, and the scene does not reveal which seat is requested. Single-modality staging never supervises such combinations, yet the parameters that will later combine vision and language keep changing. We show that this can cause lasting damage. In an analytically tractable construction, either VA or LA pretraining closes an initially active joint vision--language path. Under the specified population gradient flow, later full-input training cannot reopen it, even with unlimited adaptation, whereas the same model trained on full inputs from the start can succeed. The failure is induced by the training route, not by limited capacity or missing data. It is a counterexample, not a claim that every single-modality method must fail, but it shows that a later stage cannot be trusted to repair what an earlier stage has closed.

However, single-modality staging can damage the pathway needed to combine vision and language. For an instruction such as ``pick up the cup that is the same color as the plate,'' VA pretraining provides the scene without the instruction specifying color matching, while LA pretraining provides the instruction without the visual evidence needed to identify the matching cup. Neither stage directly supervises this joint computation, yet the parameters supporting it continue to change. The risk is therefore not merely that joint modeling is left untrained, but that its pathway can be shut down. 
% In an analytically tractable construction, either VA or LA pretraining closes an initially active joint vision--language path. 
Under the specified population gradient flow, subsequent full-input training cannot reopen it, even with unlimited adaptation. This raises a question: can action expert pretraining mitigate visual shortcuts while preserving joint vision--language modeling?

% 逻辑：指出直接 Full 训练也不是答案，提炼两项同时成立的要求，以问题句引出方法。
% Skipping staging is no remedy either: direct full-input training keeps the joint path open but leaves the visual shortcut in place. Pretraining must therefore hide visual substitutes often enough that language becomes necessary, while retaining joint evidence often enough that the computation combining both inputs is still trained. This raises a question: can one objective satisfy both requirements?

% 逻辑：从问题引出 SCAP，先说做法再说直观原理，并说明单模态与完整输入是其 mask 分布的边界情形。
% We answer with \textbf{Stochastic Context for Action Expert Pretraining (SCAP)}. SCAP randomly masks visual and text tokens before the VLM and recomputes the conditioning features from the retained inputs, keeping the action target and the always-visible action-side conditions unchanged. Masks that hide the visual substitute make the instruction necessary; masks that retain complementary cues keep supervising how vision and language are combined, which is exactly what single-modality staging withholds. As conditioning choices, VA, LA, and full-input training are deterministic boundary cases of this mask distribution.

We answer with \textbf{Stochastic Context for Action Expert Pretraining (SCAP)}. SCAP follows a two-stage training scheme: we first pretrain the action expert by randomly masking both visual and text tokens, then continue training with full visual and language inputs. By varying the visibility of both modalities, this pretraining exposes the expert to diverse combinations of visual and linguistic cues, discouraging visual shortcuts while retaining joint supervision whenever complementary cues remain visible (Fig.~\ref{fig:scap_motivation}). VA, LA, and full-input pretraining are special cases of SCAP obtained with different mask distributions.

% 逻辑：用两条理论结论对应上段的两项要求，说明 FM 附录的角色，并用一句话限定范围、引出实验。
% In the same construction, SCAP meets both requirements. First, if masks retain both joint cues sufficiently often, the joint path never closes, and the protection persists through full-input training (Sec.~\ref{sec:prevention}). Second, relative to direct full training from the same initialization, symmetric mask pretraining of any positive duration yields strictly greater language dependence and lower visual-cue-flip error after unlimited full-input adaptation when the visual branch has the larger gain, though some visual dependence remains (Sec.~\ref{sec:shortcut}). Appendix~\ref{app:fmbridge} reproduces these parameter dynamics, up to a positive time scale, within flow matching. Because these are conditional results for an explicit construction, we test whether the predicted behavior appears in a full-scale model.

In the same construction, SCAP meets both requirements. First, if masks retain both joint cues sufficiently often, the joint modeling path never closes (Sec.~\ref{sec:prevention}). Second, random masking gives cues from either modality a chance of being hidden, reducing overreliance on a single modality. This advantage persists through subsequent full-input adaptation (Sec.~\ref{sec:shortcut}).

% 逻辑：简述实际实现与两阶段训练，再按第四章三部分顺序介绍实验。

Experiments demonstrate that SCAP outperforms alternative pretraining methods, with consistent gains across four VLA architectures and multiple model scales. By intervening on test-set instructions, we further show that SCAP improves robustness to paraphrasing and sensitivity to changes in task intent. On three real-robot manipulation tasks, SCAP achieves an average success rate of 76.67\%, exceeding full-input pretraining by 25 percentage points.
% TODO（实验完成后）：在本段末尾补充一到两句主要结果；在真实数据替换第四章占位内容之前，引言不写任何实验结论。

% 逻辑：仿照 APT 以条目列出贡献，分别对应失败机制、方法与预防/缓解理论、实验评测三部分。
% In summary, our contributions are:
% \begin{itemize}[leftmargin=*,itemsep=1pt,topsep=2pt]
%     \item We identify a failure mode of single-modality action expert pretraining: in an analytical construction, VA or LA pretraining closes a joint vision--language path that full-input training cannot reopen, even with unlimited adaptation.
%     \item We propose SCAP, which masks visual and text tokens at the VLM input during expert pretraining, and prove that it prevents this closure under a joint-retention condition and, with symmetric masks, retains more language dependence than direct full training.
%     \item We evaluate SCAP against matched training routes on two simulation benchmarks, under test-time perturbations and counterfactual instructions, and on a real-robot joint-conditioning task.
% \end{itemize}

In summary, our contributions are:
\begin{itemize}[leftmargin=*,itemsep=1pt,topsep=2pt]
    \item We identify a key limitation of unimodal action expert pretraining: it can shut down certain pathways for joint vision--language modeling.
    \item We propose SCAP, which randomly masks visual and text tokens during action expert pretraining to address visual shortcuts while preserving vision--language modeling.
    \item We establish a controlled evaluation protocol spanning multi-task simulation, paired language interventions, and real-robot joint grounding to test instruction reliance and the preservation of joint conditioning.
\end{itemize}

\section{Related Work}
\label{sec:related}

% 逻辑：交代动作建模背景后平行介绍预训练路线，保留真实贡献和引用，不把条件可见性相似等同于完整算法相同。
% \paragraph{Action generation and expert pretraining.}
% Diffusion Policy~\cite{chi2023diffusionpolicy} and flow-based VLA policies such as $\pi_0$~\cite{black2024pi0} model continuous, multimodal action distributions. Several lines of work investigate how to initialize the action-generating component before full VLA adaptation. APT~\cite{xu2026apt} learns a vision--action prior and introduces language through gated feature fusion; it also analyzes visual shortcuts using a Bayesian and conditional-information perspective. LA4VLA~\cite{lin2026la4vla} constructs atomic language--action episodes and studies language-only, sequential, and mixed pretraining. Jing et al.~\cite{jing2026actionpriors} learn an action-only encoder--decoder prior, then transfer it through decoder reuse and latent distillation. ActionX~\cite{yi2026actionx} uses reinforcement learning for expert pretraining before supervised VLA adaptation. SCAP focuses on varying the visual and text evidence within each pretraining stage. Its modality-level boundary cases describe conditioning choices, not reproductions of these methods' architectures or training pipelines.

\paragraph{Action generation and expert pretraining.}
Recent VLA models, including $\pi_{0.5}$~\cite{black2025pi05} and GR00T N1.7~\cite{nvidia2026groot17}, couple pretrained VLMs with action experts for continuous robot control. Attention-based expert routing, as explored in DREAM~\cite{sheng2026dream}, addresses VLA architecture design, whereas SCAP intervenes on the conditioning context used during expert pretraining. Action expert pretraining seeks to establish useful priors before full VLA adaptation. APT~\cite{xu2026apt} and BayesVLA~\cite{xu2025bayesvla} derive staged training from a Bayesian factorization of the VLA policy into a vision--action prior and an instruction likelihood. LA4VLA~\cite{lin2026la4vla} learns instruction--action associations from atomic language--action episodes without visual input. Jing et al.~\cite{jing2026actionpriors} pretrain an action-only encoder--decoder and transfer its motion prior through decoder reuse and latent distillation. SCAP generalizes modality-restricted conditioning through simple stochastic masking to mitigate visual shortcuts while preserving joint vision--language learning.

% 逻辑：承认随机遮挡及正则化思想已有基础，定位本文研究的是专家预训练中的 VLM 输入端遮挡，而非发明 dropout。
% \paragraph{Dropout and multimodal masking.}
% Dropout regularizes neural networks by randomly removing units~\cite{srivastava2014dropout}, with established connections to adaptive regularization~\cite{wager2013dropout}. ModDrop~\cite{neverova2016moddrop} applies multimodal dropout to gesture recognition, while masked autoencoders~\cite{he2022mae} reconstruct missing image content. SCAP uses masking for a different training problem: it removes visual and text tokens \emph{at the VLM input}, recomputes the resulting multimodal representations, and supervises the action expert on the unchanged action target. The distinction from feature dropout is consequential because input masking changes the information available throughout the VLM computation. The contribution lies in this action-pretraining formulation and its analysis, not in random masking itself.

\paragraph{Dropout and multimodal masking.}
Dropout regularizes neural networks by randomly removing units~\cite{srivastava2014dropout}, with established connections to adaptive regularization~\cite{wager2013dropout}. ModDrop~\cite{neverova2016moddrop} applies multimodal dropout to gesture recognition, while masked autoencoders~\cite{he2022mae} reconstruct missing image content. SCAP uses masking to address a different training problem: it randomly masks visual and language tokens at the VLM input to mitigate visual shortcuts arising from modality imbalance in VLA pretraining data.

% 逻辑：补充与条件 dropout 的关系，说明 token 粒度与无需推理 guidance 的区别，并为 flow loss 铺垫。
% \paragraph{Condition dropout and generative policies.}
% Classifier-free diffusion guidance~\cite{ho2022cfg} learns conditional and unconditional predictions by dropping conditioning during training. SCAP shares the principle of learning under incomplete context, but samples partial subsets of both visual and text inputs and does not require conditional/unconditional score extrapolation at inference. Flow matching~\cite{lipman2023flowmatching} is one compatible action objective. We separate the conditional-learning argument from its loss-specific realization: Appendix~\ref{app:fmbridge} retains the required action-side inputs and explicitly establishes the parameter-loss correspondence, rather than claiming generic convergence or sample-quality guarantees for flow policies.

\section{Stochastic Context Pretraining}
\label{sec:method}

\begin{figure*}[t]
    \centering
    \includegraphics[width=\textwidth]{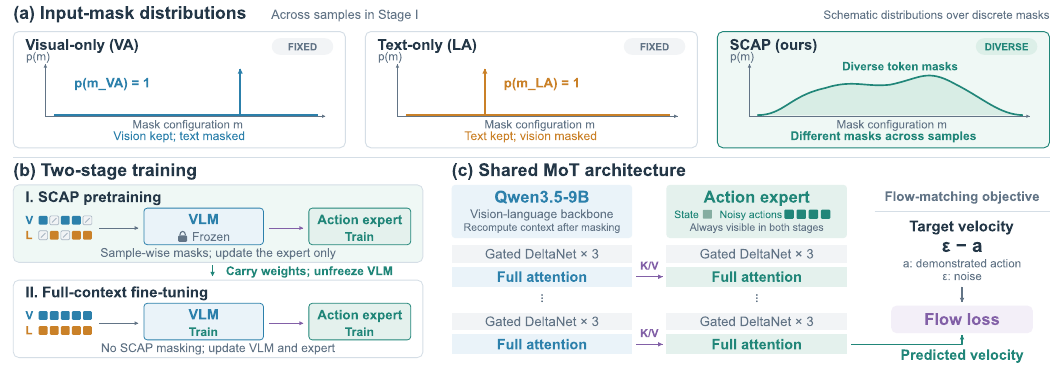}
    \caption{\textbf{SCAP training and model architecture.} (a) The action expert is pretrained with randomly masked visual and language tokens and a frozen backbone, then jointly adapted with the backbone on complete inputs. (b) The expert reads keys and values from the corresponding full-attention layers of the backbone and predicts the flow velocity.}
    \label{fig:overview}
\end{figure*}

Our question is how pretraining affects the joint action conditional $p(A\mid V,L,S)$, where $V,L,S,A$ denote vision, language, robot state, and actions. Let $Y$ denote the supervised target. With robot state $S$ always available, SCAP samples a content-independent mask $M\sim Q$ before the VLM $F_\phi$, freezes $\phi$ during expert pretraining, and optimizes the expert and its projections $\theta$:
\begin{align}
    \mathcal H_M&=F_\phi(V,L;M),\label{eq:maskedvlm}\\
    \mathcal L_{\rm mask}(\theta)&=\mathbb E_{M\sim Q}\,\ell(h_\theta(S,\mathcal H_M),Y).
    \label{eq:objective}
\end{align}
The expectation also averages over data. For theoretical analysis, we consider a simplified conditional-regression model with squared-error loss $\ell(h,Y)=\frac12\|h-Y\|^2$ (Appendix~\ref{app:diagnostic}). Appendix~\ref{app:fmbridge} shows that our simplified model and a corresponding flow-matching realization have equivalent gradient-flow dynamics up to time rescaling, so the same qualitative conclusions apply.

\subsection{Single-Modality Pretraining}
\label{sec:staging}

When visual cues alone are predictive of the target actions, the expert may rely on vision while neglecting language instructions. Single-modality staging separates the competing inputs. The vision-only (VA) and language-only (LA) pretraining objectives are:
\begin{align}
    \mathcal L_{\rm VA}&=\mathbb E\ell(h_\theta(S;V,\varnothing),Y),\nonumber\\
    \mathcal L_{\rm LA}&=\mathbb E\ell(h_\theta(S;\varnothing,L),Y).
    \label{eq:stagingobjectives}
\end{align}
Here $\varnothing$ means semantic absence: LA blocks vision, whereas VA learns visuomotor behavior before language alignment. Although staged training reduces competition between modalities, it can impair the joint vision--language modeling pathway. The next subsection proves this.

\subsection{Damage to Joint Vision--Language Modeling}
\label{sec:irreversible}

We write the visual and language inputs as $V=(Z,C)$ and $L=(D,H)$. Here, $C$ denotes a scene cue, $D$ a direct instruction cue, and $H$ a relational instruction cue; these three variables are independent and uniformly distributed over $\{-1,+1\}$. The visual cue $Z$ equals $D$ on the training distribution, providing a visual substitute for the direct instruction cue. For this diagnostic, we take robot state $S$ to be constant and define the action target as
\begin{equation}
    Y=(A_{\rm d},A_{\rm j})=(D,CH).
    \label{eq:task}
\end{equation}
Vision can substitute for language in $A_{\rm d}$ through $Z$, but $A_{\rm j}$ requires both modalities. Let $m_Z,m_C,m_D,m_H$ indicate cue presence and $j_M=m_Cm_H$. We define a simplified action predictor in which the parameter $u$ appears in both the joint-output offset and the gate controlling the vision--language interaction:
\begin{align}
    h_{\theta,\rm d}^{(M)}&=\kappa w_{\rm v}m_ZZ+w_{\rm l}m_DD,\nonumber\\
    h_{\theta,\rm j}^{(M)}&=\sqrt\lambda(b+u+c)+j_M[u]_+CH,
    \label{eq:sharedhead}
\end{align}
with trainable $(w_{\rm v},w_{\rm l},b,u)$, constants $\kappa\geq1$, $\lambda>0$, $c>1$, and $[u]_+=\max(u,0)$.

Initialize $(w_{\rm v},w_{\rm l},b,u)=(0,0,0,1)$ and write $R=b+u+c$. Let $\mathcal L_{\mathrm{single},\rm j}$ denote the expected squared-error loss for predicting the joint action $A_{\rm j}=CH$ under single-modality pretraining. For either VA or LA, $j_M=0$, giving $\mathcal L_{\mathrm{single},\rm j}=(\lambda R^2+1)/2$. With the same learning rate for $b$ and $u$, their continuous-time gradient updates are:
\begin{equation}
\begin{aligned}
    \dot b&=-\partial_b\mathcal L=-\lambda R,\\
    \dot u&=-\partial_u\mathcal L=-\lambda R,\\
    \dot R&=\dot b+\dot u=-2\lambda R.
\end{aligned}
\label{eq:singlejointflow}
\end{equation}
Here dots denote derivatives with respect to optimization time $T$. Solving $\dot R=-2\lambda R$ gives $R(T)=(1+c)e^{-2\lambda T}$. Since $\dot u=\dot b$ and $u(0)-b(0)=1$, their difference remains $u-b=1$. Combining this with $R=b+u+c$ yields
\begin{align}
    u(T)&=\tfrac{R(T)+1-c}{2}\nonumber\\
        &=\tfrac{1-c}{2}+\tfrac{1+c}{2}e^{-2\lambda T}.
    \label{eq:stageclosure}
\end{align}
Thus $u<0$ after $T_{\rm close}=\log((1+c)/(c-1))/(2\lambda)$: training closes an initially active gate. Even if the complete visual and language inputs are restored at $T_{\rm p}>T_{\rm close}$, the interaction gate remains inactive. Since $R>0$, $\dot u=-\lambda R<0$, preventing the gate from reopening. Let $T_{\rm a}\geq0$ denote the elapsed training time after restoring complete visual and language inputs. Then,
\begin{align}
    u(T_{\rm p}+T_{\rm a})&<0,\nonumber\\
    \mathcal L_{\rm full,j}&=\tfrac12(1+\lambda R^2)\geq\tfrac12,\nonumber\\
    \lim_{T_{\rm a}\to\infty}\mathcal L_{\rm full,j}&=\tfrac12>0.
    \label{eq:irreversible}
\end{align}
Thus, in this model, single-modality pretraining can close the joint vision--language pathway, and subsequent training with complete visual and language inputs cannot reopen it under the same gradient-flow dynamics. The detailed derivation is provided in Appendix~\ref{app:closure}.

\subsection{Preserving Joint Modeling with SCAP}
\label{sec:prevention}

The diagnostic assumes independent retention of visual and language cues with probability $q\in(0,1)$. The probability of retaining both cues required for joint prediction is therefore $\alpha=\Pr(j_M=1)=q^2$. Averaging over all masks gives
\begin{align}
    \mathcal L_{\mathrm{mask},\rm j}&=\tfrac{1-\alpha}{2}+F_\alpha(b,u),\nonumber\\
    F_\alpha(b,u)&=\tfrac12\left[\lambda(b+u+c)^2
                              +\alpha(1-[u]_+)^2\right].
    \label{eq:maskenergy}
\end{align}
Appendix~\ref{app:prevention} derives this expectation. Missing-cue masks still contribute offset loss. For $u>0$, $\dot u=-\lambda R+\alpha(1-u)$: joint evidence opposes closure. If $\lambda(1+c)^2<\alpha$, then
\begin{align}
    F_\alpha(0,1)&<\tfrac\alpha2\leq F_\alpha(b,0),\quad\forall b\in\mathbb R,\nonumber\\
    \frac{\mathrm dF_\alpha}{\mathrm dT}
       &=-\|\nabla_{b,u}F_\alpha\|^2\leq0.
    \label{eq:barrier}
\end{align}
Since gradient flow cannot increase $F_\alpha$, the trajectory cannot reach $u=0$, and the joint pathway remains active. However, this guarantee requires a sufficiently high retention probability $q$.

In the second stage, we remove the masks and continue training with complete visual and language inputs. At the switch, $F_1\leq F_\alpha/\alpha<1/2$, so the full-input joint loss is below the minimum loss at the gate-closing boundary $u=0$. Gradient flow therefore keeps the joint pathway active and converges to zero joint prediction error in our simplified model. The detailed derivation is provided in Appendix~\ref{app:prevention}.

Beyond preserving the joint pathway, we next show that SCAP reduces reliance on visual shortcuts compared with direct full-input training.

\subsection{Reducing Visual Shortcuts with SCAP}
\label{sec:shortcut}

We use the subscripts $\mathrm{F}$ and $\mathrm{MF}$ to denote direct full-input training and SCAP pretraining followed by full-input training, respectively. Both routes start from the same initialization: F uses complete inputs for $T_{\rm p}+T_{\rm a}$, whereas MF uses masked inputs for $T_{\rm p}>0$ and then complete inputs for $T_{\rm a}$. We use the same masking scheme with retention probability $q$ and assume $\kappa>1$; $\lambda(1+c)^2<q^2$ simultaneously protects the joint path.

Write $a=\kappa w_{\rm v}$, $d=w_{\rm l}$, $g=\kappa^2>1$, and $\beta=g+1$.

\paragraph{Visual Dominance under Full-Input Training.}
For the direct action component, the full-input loss is
\begin{equation}
\begin{aligned}
    \mathcal L_{\mathrm{full},\rm d}
    &=\frac12\mathbb E\!\left[
    \left(h_{\theta,\rm d}^{(\mathrm{full})}-A_{\rm d}\right)^2
    \right]\\
    &=\frac12(a+d-1)^2.
\end{aligned}
\label{eq:fulldirectloss}
\end{equation}
Here $h_{\theta,\rm d}^{(\mathrm{full})}=aZ+dD$ denotes the direct action prediction with complete inputs, and $Z=D=A_{\rm d}$ on the training distribution. Its gradient-flow dynamics are:
\begin{equation}
\begin{alignedat}{2}
    &\dot a=g(1-a-d),\qquad &&\dot d=1-a-d,\\
    &\tfrac{\mathrm d}{\mathrm dT}(a-gd)=0,\qquad &&a(0)=d(0)=0.
\end{alignedat}
\label{eq:fullrouteodes}
\end{equation}
The invariant $a=gd$ yields
\begin{align}
    a_{\rm F}(T)&=\tfrac g\beta(1-\exp(-\beta T)),\nonumber\\
    d_{\rm F}(T)&=\tfrac1\beta(1-\exp(-\beta T)).
    \label{eq:fullroutesolution}
\end{align}
The limiting coefficients are $a_{\rm F}^{\infty}=g/(g+1)$ and $d_{\rm F}^{\infty}=1/(g+1)$. As $T\to\infty$, the training error vanishes, but vision dominates when $g>1$. Thus, direct full-input training can favor visual cues over language instructions. We next show that SCAP pretraining mitigates this visual shortcut.

\paragraph{Language-Dominant Solutions with SCAP.}
Let $(a_{\rm p},d_{\rm p})$ be the coefficients actually reached after a finite SCAP pretraining duration $T_{\rm p}>0$, without assuming convergence of that stage. Continuing with complete visual and language inputs and taking $T_{\rm a}\to\infty$ gives
\begin{equation}
\begin{aligned}
    a_{\rm MF}^{\infty}&=\tfrac{g+a_{\rm p}-gd_{\rm p}}\beta,\\
    d_{\rm MF}^{\infty}&=\tfrac{1+gd_{\rm p}-a_{\rm p}}\beta.
\end{aligned}
\label{eq:fullendpointlimits}
\end{equation}
These limits follow because $a+d\to1$ while $gd-a$ remains constant during full-input training. Relative to F,
\begin{equation}
\begin{aligned}
    d_{\rm MF}^{\infty}-d_{\rm F}^{\infty}
      &=\tfrac{gd_{\rm p}-a_{\rm p}}\beta,\\
    a_{\rm MF}^{\infty}-a_{\rm F}^{\infty}
      &=-\tfrac{gd_{\rm p}-a_{\rm p}}\beta.
\end{aligned}
\label{eq:endpointcomparison}
\end{equation}
The sign of $gd_{\rm p}-a_{\rm p}$ determines whether MF increases language reliance and reduces visual reliance relative to F. We therefore define
\begin{equation}
\begin{aligned}
    \delta(T)&:=gd(T)-a(T),\\
    \delta_{\rm p}&:=\delta(T_{\rm p})=gd_{\rm p}-a_{\rm p}.
\end{aligned}
\label{eq:deltadefinition}
\end{equation}
Thus $\delta_{\rm p}>0$ would establish increased language reliance and reduced visual reliance. To prove its sign along the actual pretraining trajectory, average the masked direct-coordinate error:
\begin{align}
    \mathcal L_{\mathrm{mask},\rm d}
      &=\tfrac12[qa^2+qd^2+2q^2ad\nonumber\\
      &\qquad-2q(a+d)+1].
    \label{eq:maskdirectloss}
\end{align}
The corresponding dynamics are
\begin{align}
    \dot a&=gq(1-a-qd),\nonumber\\
    \dot d&=q(1-d-qa),\qquad a(0)=d(0)=0.
    \label{eq:maskdirectodes}
\end{align}
Along mask training, $\dot\delta=gq(1-q)(a-d)$ and $\delta(0)=0$. The exact solution for $a-d$ is a positive multiple of the difference between a slower and a faster decaying exponential, so $a(T)>d(T)$ for every $T>0$ (Appendix~\ref{app:direct}). Integrating gives $\delta_{\rm p}>0$ for every finite $T_{\rm p}>0$. Thus, MF retains greater language dependence than F.

Language dominance, $d_{\rm MF}^{\infty}>a_{\rm MF}^{\infty}$, further requires $\delta_{\rm p}>(g-1)/2$. As $T_{\rm p}\to\infty$, the masked coefficients satisfy $a_{\rm p},d_{\rm p}\to1/(1+q)$, giving $\delta_{\rm p}\to(g-1)/(1+q)>(g-1)/2$. Thus sufficiently long but finite SCAP pretraining yields language dominance after full-input training.

\paragraph{Lower Loss under Visual-Cue Flips.}
To assess reliance on the visual shortcut, we make the shortcut cue conflict with the language instruction. Specifically, we flip $Z=D$ to $Z=-D$, keeping the instruction, target, and other cues unchanged. The direct-coordinate loss becomes
\begin{align}
    \mathcal E_{\rm flip}(a,d)
      &=\tfrac12\mathbb E[((d-a)D-D)^2]\nonumber\\
      &=\tfrac12(1+a-d)^2.
    \label{eq:flipmetric}
\end{align}
At full-input convergence, the prediction $aZ+dD=(a+d)D$ on the training distribution matches the target $D$, giving $a+d=1$. Substituting $d=1-a$ into the flip loss yields $\mathcal E_{\rm flip}=2a^2$. Since $0<a_{\rm MF}^{\infty}<a_{\rm F}^{\infty}$, the flip losses satisfy
\begin{align}
    \mathcal E_{\rm flip,MF}^{\infty}
       &=2(a_{\rm MF}^{\infty})^2\nonumber\\
       &<2(a_{\rm F}^{\infty})^2
        =\mathcal E_{\rm flip,F}^{\infty}.
    \label{eq:shortcutcomparison}
\end{align}
This loss advantage holds after full-input convergence for every finite $T_{\rm p}>0$. Thus, compared with direct full-input training, SCAP enables more robust instruction following under this visual-cue perturbation. Appendix~\ref{app:direct} provides the detailed trajectories, finite-time comparisons, and an asymmetric-retention extension.

\subsection{Architecture and Two-Stage Training}
\label{sec:implementation}

\paragraph{Model configuration.}
Our configuration uses a mixture-of-transformers (MoT) architecture with a Qwen3.5-9B vision--language backbone~\cite{qwen2026qwen35nine} and an approximately 900M-parameter action expert. Figure~\ref{fig:overview} illustrates the two-stage training process and the overall model architecture.

\paragraph{Stage 1: masked context, expert-only updates.}
We initialize and freeze the pretrained VLM. After visual encoding and before the language backbone, SCAP uniformly selects a fixed-size subset (30\%) of pooled image-content and instruction tokens and zeros their complete embeddings in place. Token positions are preserved, with no mask-token replacement or rescaling. Only the expert and action-side projections are optimized using Eq.~\eqref{eq:objective}; Appendix~\ref{app:token-masking} gives implementation details.

\paragraph{Stage 2: complete context, joint model updates.}
We carry the Stage-1 weights into full-context VLA training. SCAP input masking is disabled, and the VLM is unfrozen; both $\phi$ and $\theta$ are updated on complete visual observations and instructions using the same action-target convention. Inference uses complete, unmasked inputs.

\section{Experiments}
\label{sec:experiments}

We evaluate SCAP through multi-task performance, language robustness and intent sensitivity, and language-conditioned real-robot manipulation.

\begin{table*}[t]
    \centering
    \small
    \setlength{\tabcolsep}{5.5pt}
    \begin{tabular}{l|ccc|ccc|c|c}
      \hline
      \multicolumn{1}{c|}{\multirow{2}{*}{Method}}
      & \multicolumn{3}{c|}{RoboTwin 2.0}
      & \multicolumn{3}{c|}{RoboCasa-GR1}
      & \multicolumn{1}{c|}{\multirow{2}{*}{Avg.}}
      & \multirow{2}{*}{Gain} \\
      \cline{2-4}\cline{5-7}
      & \multicolumn{1}{c}{Clean} & \multicolumn{1}{c}{Randomized} & \multicolumn{1}{c|}{Overall}
      & \multicolumn{1}{c}{Articulated} & \multicolumn{1}{c}{Pick-and-Place} & \multicolumn{1}{c|}{Overall} & & \\
      \hline
        % $\pi_0$~\cite{black2024pi0} & 38.92 & 36.20 & 37.56 & - & - & - & - & - \\
        GO-1~\cite{bu2025agibot_iros} & 37.80 & 36.20 & 37.00 & 35.73\rlap{$^{*}$} & 43.13\rlap{$^{*}$} & 41.28\rlap{$^{*}$} & 39.14\rlap{$^{*}$} & - \\
        $\pi_{0.5}$~\cite{black2025pi05} & 42.98 & 43.84 & 43.41 & 44.24\rlap{$^{*}$} & 48.67\rlap{$^{*}$} & 47.54\rlap{$^{*}$} & 45.48\rlap{$^{*}$} & - \\
        GR00T N1.6~\cite{nvidia2026gr00tn16eval} & 47.34\rlap{$^{*}$} & 44.98\rlap{$^{*}$} & 46.16\rlap{$^{*}$} & 24.17 & \underline{55.43} & 47.60 & 46.88\rlap{$^{*}$} & - \\
        RDT-1B~\cite{liu2024rdt} & 45.15\rlap{$^{*}$} & 43.43\rlap{$^{*}$} & 44.29\rlap{$^{*}$} & 40.05\rlap{$^{*}$} & 46.61\rlap{$^{*}$} & 44.97\rlap{$^{*}$} & 44.63\rlap{$^{*}$} & - \\
        OpenVLA-OFT~\cite{kim2024openvla,kim2025oft} & \textbf{53.63}\rlap{$^{*}$} & \underline{50.85}\rlap{$^{*}$} & \textbf{52.24}\rlap{$^{*}$} & \underline{46.51}\rlap{$^{*}$} & 51.27\rlap{$^{*}$} & \underline{50.08}\rlap{$^{*}$} & \underline{51.16}\rlap{$^{*}$} & - \\
        % FLARE Table 1 reports Art.=51.3, PnP=58.2, Overall=55.0.
        % Preserve all three scores as reported in FLARE Table 1; do not recompute Overall.
        FLARE~\cite{zheng2025flare} & \underline{52.50}\rlap{$^{*}$} & \textbf{51.04}\rlap{$^{*}$} & \underline{51.77}\rlap{$^{*}$} & \textbf{51.30} & \textbf{58.20} & \textbf{55.00} & \textbf{53.39}\rlap{$^{*}$} & - \\
      \hline
        Qwen3.5-9B~\cite{qwen2026qwen35nine} & 34.40 & 33.40 & 33.90 & 40.42 & 48.03 & 46.13 & 40.02 & +0.00 \\
        \quad Full & 42.08 & 42.04 & 42.06 & 42.46 & 48.51 & 47.00 & 44.53 & +4.51 \\
        \quad LA$\rightarrow$Full (LA4VLA~\cite{lin2026la4vla}) & 43.40 & 42.12 & 42.76 & 46.49 & 46.61 & 46.58 & 44.67 & +4.65 \\
        \quad VA$\rightarrow$Full (APT~\cite{xu2026apt}) & 50.44 & 49.88 & 50.16 & 47.18 & 47.50 & 47.42 & 48.79 & +8.77 \\
        \quad VL-Mask (50\%)$\rightarrow$Full & 46.12 & 46.48 & 46.30 & 44.93 & 50.25 & 48.92 & 47.61 & +7.59 \\
      \hline
        \rowcolor[RGB]{244,250,248}
        \quad \textbf{SCAP}$\rightarrow$Full (freeze VLM) & \underline{53.50} & \underline{52.42} & \underline{52.96} & \underline{54.08} & \underline{53.44} & \underline{53.60} & \underline{53.28} & \underline{+13.26} \\
        \rowcolor[RGB]{244,250,248}
        \quad \textbf{SCAP}$\rightarrow$Full & \textbf{56.66} & \textbf{56.30} & \textbf{56.48} & \textbf{59.53} & \textbf{58.29} & \textbf{58.60} & \textbf{57.54} & \textbf{+17.52} \\
      \hline
    \end{tabular}
    \caption{\textbf{Results on RoboTwin 2.0 and RoboCasa-GR1 (success rate \%).} FLARE's RoboCasa-GR1 scores are quoted directly from Table~1 of the original paper; its reported Overall score is not recomputed from the subgroup scores. Indented routes are trained independently from the same backbone initialization; $\rightarrow$ denotes stage order within each route. $^{*}$ marks scores from reproductions aligned with our training configuration. \textbf{Best} and \underline{second-best} scores are bold and underlined, respectively.}
    \label{tab:simulator-main}
\end{table*}

\begin{table*}[!t]
    \centering
    \begin{minipage}[t]{0.48\textwidth}
        \vspace{0pt}
        \centering
        \small
        \begingroup
        \setlength{\tabcolsep}{3.5pt}
        \renewcommand{\arraystretch}{1.08}
        \begin{tabular}{@{}lcc|cc|c@{}}
            \hline
            Method & OS $\uparrow$ & PS $\uparrow$ & OG $\downarrow$ & NG $\uparrow$ & PC $\uparrow$ \\
            \hline
            Qwen3.5-9B~\cite{qwen2026qwen35nine} & 53.37 & 48.91 & 34.40 & 25.49 & 19.58 \\
            Full & 54.03 & 50.63 & 32.09 & 27.26 & 21.45 \\
            LA$\rightarrow$Full & 54.90 & 52.14 & \underline{22.84} & 35.31 & 28.97 \\
            VA$\rightarrow$Full & 57.19 & 52.92 & 35.17 & 31.07 & 25.36 \\
            VL-Mask (50\%)$\rightarrow$Full & \underline{59.26} & \underline{56.05} & 23.13 & \underline{38.90} & \underline{32.78} \\
            \textbf{SCAP}$\rightarrow$Full & \textbf{67.23} & \textbf{65.83} & \textbf{12.24} & \textbf{56.46} & \textbf{47.93} \\
            \hline
        \end{tabular}
        \endgroup
        \captionof{table}{\textbf{Text perturbations on RoboCasa-GR1 sub-tasks (\%).} OS: Original SR; PS: Paraphrase SR; OG: Old-Goal rate; NG: New-Goal SR; PC: Pair-Correct rate. \textbf{Best} and \underline{second-best} scores are bold and underlined, respectively.}
        \label{tab:instruction-interventions}
    \end{minipage}\hfill
    \begin{minipage}[t]{0.48\textwidth}
        \vspace{0pt}
        \centering
        \includegraphics[width=\linewidth,trim=0 8pt 0 18pt,clip]{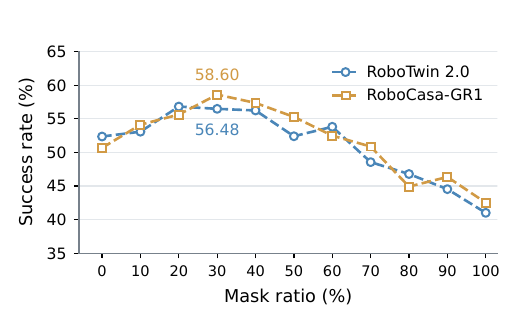}
        \captionsetup{skip=4pt}
        \captionof{figure}{\textbf{Token Mask-ratio ablation.} Overall success rates on RoboTwin 2.0 and RoboCasa-GR1.}
        \label{fig:mask-ratio}
    \end{minipage}
    \par\vspace{-8pt}
\end{table*}

\subsection{Experimental Setup}
\label{sec:experimental-setup}

\paragraph{Benchmarks.}
We use two simulation benchmarks. RoboTwin 2.0~\cite{chen2025robotwin2} comprises 50 bimanual manipulation tasks with Clean (Easy) and Randomized (Hard) evaluation settings. RoboCasa-GR1~\cite{nvidia2025gr00tn1,robocasaGR1Tabletop} contains 24 humanoid tabletop tasks, including 6 PnPClose and 18 PosttrainPnP tasks. We report success rates for each setting or group and the within-benchmark average.

\paragraph{Implementation Details.}
Both pretraining stages are trained for 100k optimization steps with a global batch size of 512, while benchmark-specific SFT uses a global batch size of 256. We adopt AdamW with a weight decay of 0.01 and an initial action-expert learning rate of $10^{-4}$. A 1000-step warmup is followed by cosine learning-rate decay, with a minimum VLM learning rate of $10^{-6}$ in Stage 2. Visual observations comprise up to three camera views at $224\times224$ resolution, and the policy predicts 16-step action chunks. 
% All training is performed in BF16 precision.

\subsection{Training Data}
\label{sec:training-data}

We use approximately 1.2 million action-labeled trajectories from real-robot and synthetic datasets, processed with a shared pipeline. State/action fields share a 46-dimensional layout with quaternion end-effector poses and calibrated gripper commands, while retaining native coordinate frames and reference points. Quality checks assess pose availability, kinematic consistency, temporal smoothness, state--action alignment, and numerical validity. Controlled routes share data within each stage; instruction interventions are evaluation-only. Appendix~\ref{app:training-data} details source composition, conversion rules, filtering statistics, and reporting limitations.

\subsection{Main Results}
\label{sec:simulator-main}

\paragraph{RoboTwin 2.0.}
Table~\ref{tab:simulator-main} shows that SCAP achieves the highest success rates in both Clean and Randomized settings, reaching 56.48\% overall success. It outperforms Full, VA$\rightarrow$Full, and VL-Mask$\rightarrow$Full by 14.42, 6.32, and 10.18 percentage points, respectively. These comparisons support stochastic joint-context pretraining over vision-only conditioning or complete-modality dropout. Joint Stage-2 fine-tuning further improves overall success from 52.96\% with a frozen VLM to 56.48\%, highlighting the benefit of backbone adaptation after action-expert pretraining.

\paragraph{RoboCasa-GR1.}
SCAP also achieves the best results in both Articulated and Pick-and-Place categories, with 58.60\% overall success. It surpasses Full, VA$\rightarrow$Full, and VL-Mask$\rightarrow$Full by 11.60, 11.18, and 9.68 percentage points, respectively, in overall success rate. Joint Stage-2 fine-tuning improves overall success from 53.60\% to 58.60\%, extending the benefit of backbone adaptation to humanoid tabletop manipulation. The consistent gains across both benchmarks show the effectiveness of SCAP pretraining followed by full-context adaptation.

\subsection{Ablation Studies}
\label{sec:ablations}

\paragraph{Text Perturbations.}
\label{sec:test-interventions}
\label{sec:instruction-interventions}
Guided by the visual-shortcut analysis in~\cref{sec:shortcut}, we compare Qwen3.5-9B and five training routes on four RoboCasa-GR1 pick-and-place tasks~\cite{robocasaGR1Tabletop}. Paraphrases preserve intent; minimal object or destination edits change the goal in the same scene, probing cue--instruction conflict rather than directly flipping visual cues. Each task uses 100 shared scenes and two paraphrases per scene, with independently reset rollouts, complete inputs, and matched inference settings. Table~\ref{tab:instruction-interventions} reports equal-weight task means. SCAP's high paraphrase success and small drop from original instructions indicate wording robustness while retaining semantic sensitivity. Lower old-goal execution and higher new-goal success indicate redirection to the requested goal rather than inaction. Paired correctness requires both original- and edited-instruction rollouts to succeed and cannot exceed either single-instruction success rate. It combines goal selection and physical execution, whose residual failures cannot be separated by aggregate scores. Together, these results support reduced visual shortcuts and preserved joint vision--language grounding after full-context adaptation.

\begin{figure*}[!t]
    \centering
    \includegraphics[width=\textwidth]{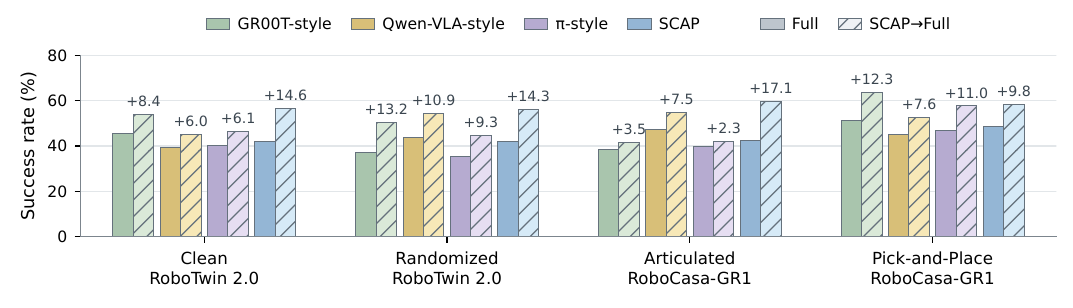}
    \caption{\textbf{Action expert pretraining across VLA architectures.}}
    \label{fig:cross-architecture}
\end{figure*}

\paragraph{Token Masking Ratio.}
We vary Stage-1 masking over $r\in\{0,0.1,\ldots,1\}$ with fixed data, stage budgets, and evaluation settings. Stage 2 and inference use complete inputs; $r=0$ gives two full-context stages on the same data (Full$\rightarrow$Full). In Figure~\ref{fig:mask-ratio}, RoboTwin 2.0 peaks at 20\% and RoboCasa-GR1 at 30\%; we adopt 30\% for its highest cross-benchmark average. The 20--40\% range remains competitive, while stronger masking shows an overall decline with local rebounds. SCAP's advantage over the budget-matched zero-mask control cannot be explained solely by an extra round of training on the same data. This agrees with the qualitative trade-off in Eq.~\eqref{eq:barrier}: $\lambda(1+c)^2<q^2$ is sufficient to preserve the joint pathway. Moderate masking weakens shortcuts while retaining task-critical joint evidence. Here $q$ denotes semantic-cue retention, not token retention, so the diagnostic does not predict an optimal token ratio.

\paragraph{Action Pretraining on More Architectures.}
We test final-feature cross-attention (GR00T-style), concatenated-feature DiT (Qwen-VLA-style~\cite{wang2026qwenvla}), layer-wise full-attention MoT ($\pi$-style~\cite{black2024pi0,hancock2026gradientrouting}), and hybrid-attention MoT (SCAP). The $\pi$-style uses Qwen3-VL-8B, the nearest dense scale; the others use Qwen3.5-9B. Figure~\ref{fig:cross-architecture} compares Full with SCAP$\rightarrow$Full under matched downstream SFT and inference settings within each architecture. Gains across all four benchmark settings, despite varying magnitudes, support transfer across representative VLA conditioning interfaces rather than dependence on one fusion design.

\paragraph{Model Scale.}
We test Qwen3.5 at 2B, 4B, and 9B with approximately 100M, 300M, and 900M action experts, respectively. Within each scale, Full and SCAP$\rightarrow$Full share the backbone, expert, pretraining corpus, and downstream SFT settings. Consistent gains on both benchmarks in Table~\ref{tab:model-scale} indicate that SCAP benefits compact and larger models as backbone and expert capacity vary. Same-scale reproductions provide context rather than matched controls because their expert capacities and prior pretraining differ.

\begin{table}[t]
    \centering
    \small
    \begingroup
    \setlength{\tabcolsep}{2.5pt}
    \renewcommand{\arraystretch}{1.08}
    \begin{tabular}{@{}cclcc@{}}
      \hline
      \multicolumn{2}{c}{Model Size} & \multirow{2}{*}{Method} & \multirow{2}{*}{RoboTwin} & \multirow{2}{*}{GR1} \\
      \cline{1-2}
      VLM & Expert & & & \\
      \hline
      \multirow{3}{*}{2B} & $\sim$100M & Full & 34.86 & 36.73 \\
      & $\sim$100M & APT~\cite{xu2026apt} & \underline{39.15}\rlap{$^{*}$} & \underline{41.95}\rlap{$^{*}$}  \\
      & $\sim$100M & \textbf{SCAP}$\rightarrow$Full & \textbf{44.45} & \textbf{47.54} \\
      \hline
      \multirow{4}{*}{4B} & $\sim$300M & Full & 40.12 & 42.97 \\
      & $\sim$160M & ABot-M0~\cite{yang2026abotm0} & 46.39\rlap{$^{*}$} & 49.06\rlap{$^{*}$}  \\
      & $\sim$1.15B & Qwen-VLA-style~\cite{wang2026qwenvla} & \underline{50.67}\rlap{$^{*}$} & \textbf{55.72}\rlap{$^{*}$}  \\
      & $\sim$300M & \textbf{SCAP}$\rightarrow$Full & \textbf{51.48} & \underline{54.58} \\
      \hline
      \multirow{2}{*}{9B} & \multirow{2}{*}{$\sim$900M} & Full & \underline{42.06} & \underline{47.00} \\
      & & \textbf{SCAP}$\rightarrow$Full & \textbf{56.48} & \textbf{58.60} \\
      \hline
    \end{tabular}
    \endgroup
    \caption{\textbf{Model-scale ablation.}  $^{*}$ marks scores from reproductions aligned with our training configuration. \textbf{Best} and \underline{second-best} scores within each scale are bold and underlined, respectively.}
    \label{tab:model-scale}
\end{table}

\begin{figure*}[!t]
    \centering
    \includegraphics[width=\textwidth]{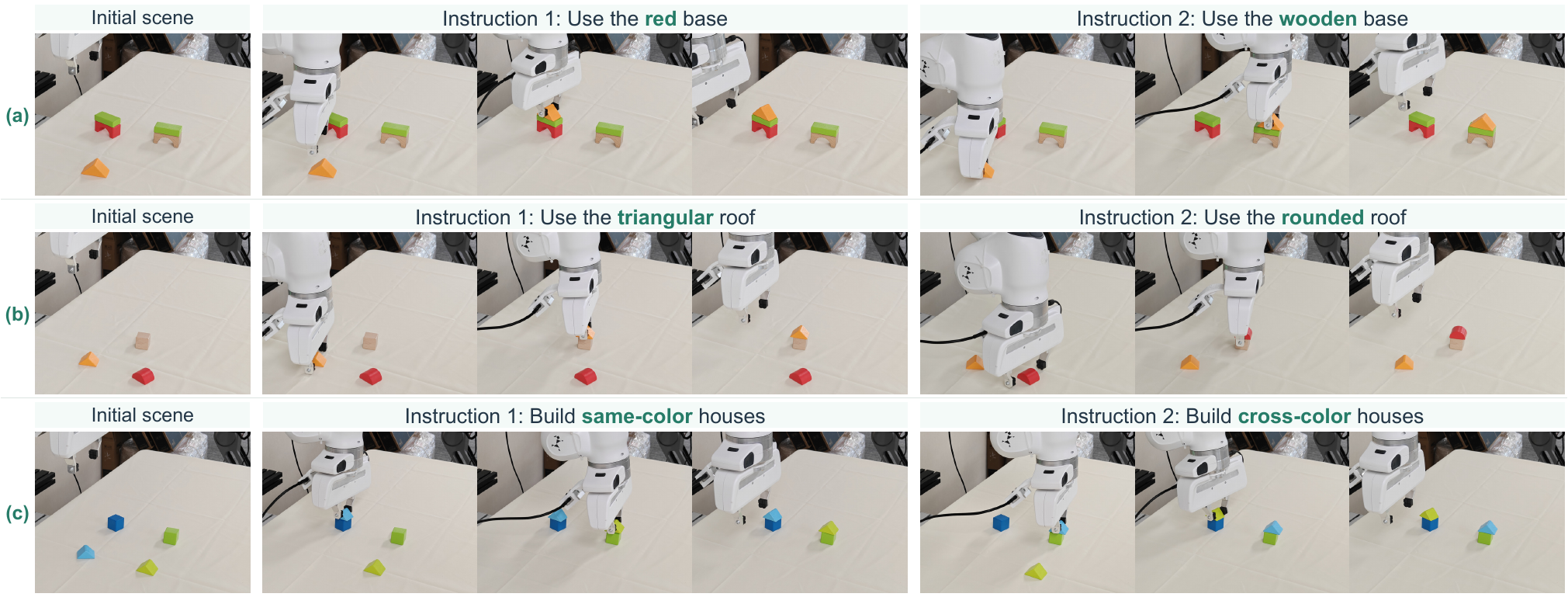}
    \caption{\textbf{Real-robot roof-assembly sequences under different language instructions.} Each row shows a shared initial scene followed by two instruction-conditioned execution sequences, read left to right within each group. (a) Place the orange roof on the red or natural-wood base. (b) Install the triangular or rounded roof. (c) Assemble both houses with same-color or cross-color roof--body pairings. Instruction keywords are highlighted. Photographs illustrate task execution rather than aggregate policy performance.}
    \label{fig:robot-frame-template}
\end{figure*}

\begin{table*}[!t]
\centering
\small
\setlength{\tabcolsep}{4.7pt}
\begin{tabular}{@{}l|ccc|ccc|ccc|c@{}}
\hline
\multicolumn{1}{c|}{\multirow{2}{*}{Method}} & \multicolumn{3}{c|}{Base selection} & \multicolumn{3}{c|}{Roof selection} & \multicolumn{3}{c|}{Color matching} & \multicolumn{1}{c}{\multirow{2}{*}{Overall}} \\
\cline{2-4}\cline{5-7}\cline{8-10}
 & \multicolumn{1}{c}{Red} & \multicolumn{1}{c}{Wood} & \multicolumn{1}{c|}{Average} & \multicolumn{1}{c}{Triangular} & \multicolumn{1}{c}{Rounded} & \multicolumn{1}{c|}{Average} & \multicolumn{1}{c}{Same} & \multicolumn{1}{c}{Cross} & \multicolumn{1}{c|}{Average} & \\
\hline
Qwen3.5-9B~\cite{qwen2026qwen35nine} & 50 & 40 & 45 & 40 & 60 & 50 & 40 & 20 & 30 & 41.67 \\
\quad Full & 40 & 60 & 50 & 50 & 70 & 60 & 40 & 50 & 45 & 51.67 \\
\quad LA$\rightarrow$Full (LA4VLA~\cite{lin2026la4vla}) & \underline{70} & 30 & 50 & 60 & 40 & 50 & 40 & 30 & 35 & 45.00 \\
\quad VA$\rightarrow$Full (APT~\cite{xu2026apt}) & \underline{70} & \underline{70} & 70 & 70 & 60 & 65 & \textbf{70} & 40 & \underline{55} & 63.33 \\
\quad VL-Mask (50\%)$\rightarrow$Full & 50 & \textbf{80} & 65 & 60 & \underline{80} & \underline{70} & 50 & 30 & 40 & 58.33 \\
\hline
\rowcolor[RGB]{244,250,248}
\quad \textbf{SCAP}$\rightarrow$Full (freeze VLM) & \textbf{80} & \underline{70} & \underline{75} & \underline{80} & \textbf{90} & \textbf{85} & 50 & \underline{60} & \underline{55} & \underline{71.67} \\
\rowcolor[RGB]{244,250,248}
\quad \textbf{SCAP}$\rightarrow$Full & \textbf{80} & \textbf{80} & \textbf{80} & \textbf{90} & \underline{80} & \textbf{85} & \underline{60} & \textbf{70} & \textbf{65} & \textbf{76.67} \\
\hline
\end{tabular}
\caption{\textbf{Real-robot success rates (\%).} Indented routes are trained independently from the same backbone initialization; $\rightarrow$ denotes stage order within each route. Each instruction has 10 trials; Average is the mean of the two instruction scores; Overall averages the three task scores. \textbf{Best} and \underline{second-best} scores in each column are bold and underlined, respectively; ties receive the same formatting. See Appendix~\ref{app:real-robot-details} for protocol and initialization details.}
\label{tab:robot-tabletop-results}
\end{table*}

\subsection{Real-Robot Evaluation}
\label{sec:real-robot}

\paragraph{Tasks.}
Three wooden-block tasks require different goals under two language instructions for the same initial scene (Figure~\ref{fig:robot-frame-template}). \emph{Base selection} places an orange triangular roof on the red or natural-wood base, specified by color. \emph{Roof selection} installs an orange triangular or red rounded roof; the candidates differ in both shape and color. \emph{Color-relation matching} assembles two houses from blue and green bodies and roofs using same-color or cross-color pairings.

\paragraph{Data and adaptation.}
We collect 50 demonstrations per instruction (300 total) and adapt the seven simulation initialization routes independently to each task. All routes share task-specific data, preprocessing, normalization, and adaptation budgets. All real-robot SFT freezes the VLM and updates the action expert and state/action projections. Inputs comprise three RGB views, joint angles, and gripper opening; targets are the next retained joint and gripper state after idle removal. We use chunk length 32, batch size 64, AdamW learning rate $5\times10^{-5}$, and the final 10,000-update checkpoint. Appendix~\ref{app:real-robot-details} provides implementation details.

\paragraph{Evaluation protocol.}
Each task-specific policy is tested on the same 10 held-out layouts under both instructions (20 episodes per task, 420 total). The scene, policy history, and action cache are reset before each instruction, with randomized instruction order. Success requires fulfilling the instruction and keeping the blocks from falling after the gripper releases them. Table~\ref{tab:robot-tabletop-results} reports instruction-wise SR and their mean to expose target preferences.

\paragraph{Results.}
SCAP$\rightarrow$Full achieves mean success rates of 80\%, 85\%, and 65\% on base selection, roof selection, and color matching, respectively, outperforming VA$\rightarrow$Full (70\%, 65\%, and 55\%) and LA$\rightarrow$Full (50\%, 50\%, and 35\%). Beyond identifying a single target, color matching requires applying a language-specified relation to multiple visual objects across two assembly operations. Because the same scene supports either pairing, visual layout alone cannot determine the requested configuration. On same-/cross-color assembly, SCAP$\rightarrow$Full achieves 60\%/70\%, compared with 70\%/40\% for VA$\rightarrow$Full and 40\%/30\% for LA$\rightarrow$Full. Its advantage over both unimodal routes is largest in cross-color assembly, which requires pairing objects according to the instruction rather than color similarity. Under shared adaptation data and budgets, these results provide behavioral support for SCAP better preserving the capacity for subsequent joint vision--language learning.

\section{Conclusion}
\label{sec:conclusion}

% 逻辑：按四节顺序总结，保持比有限效率损失更强的结论，同时把损失实现与条件依赖学习分开表述。
We show that single-modality action-expert pretraining can compromise joint vision–language modeling. To address this, we introduce SCAP, a pretraining stage that randomly masks visual and text tokens before full-input training. Our diagnostic analysis shows that SCAP can preserve joint modeling pathways while reducing visual shortcuts and strengthening reliance on language instructions. Experiments across different VLA architectures and real-robot manipulation tasks demonstrate consistent performance improvements. More broadly, SCAP provides a unified framework for action-expert pretraining based on mask distributions, with vision-only, language-only, and full-input pretraining recovered as special cases.

\section*{LLM Usage Statement}
Large language models were used to assist with writing and revising code, including figure-preparation scripts, and with language editing and grammar polishing of the manuscript. The authors take full responsibility for the code, experimental results, and final manuscript.

{
    \small
    \bibliographystyle{ieeenat_fullname}
    \bibliography{main}
}

\clearpage
\appendix

\noindent\textbf{Organization.} Appendices~\ref{app:diagnostic}--\ref{app:direct} provide the methodological derivations, followed by the token-masking protocol (Appendix~\ref{app:token-masking}), in the order of Sec.~\ref{sec:method}. The experimental appendices follow Sec.~\ref{sec:experiments}: training data and preprocessing (Appendix~\ref{app:training-data}) and real-robot evaluation (Appendix~\ref{app:real-robot-details}).
\section{Construction of the Diagnostic Model}
\label{app:diagnostic}

% 逻辑：一次明确全部变量与结构限制，强调背景常数只属于这个诊断，不是对真实专家动作侧输入的假设。
To analyze how pretraining input visibility affects joint vision--language modeling and reliance on visual shortcuts, we construct a simplified conditional-regression diagnostic model. The independent fair signs $C,D,H$ represent scene, direct instruction, and relational instruction cues. Set $Z=D$, $V=(Z,C)$, $L=(D,H)$, and $Y=(D,CH)$. For this diagnostic, robot state $S$ is constant. The head in Eq.~\eqref{eq:sharedhead} has trainable parameters $(w_{\rm v},w_{\rm l},b,u)$ and fixed constants $\kappa\geq1$, $\lambda>0$, $c>1$.

\subsection{A General Loss Formulation}

% 逻辑：先写预测残差，随后利用零均值与二值 mask，逐项得到两块参数独立的统一目标。
The goal of this step is to obtain one objective for all visibility routes and verify that the direct and joint parameter blocks can be analyzed separately. Let $R=b+u+c$ and $a=\kappa w_{\rm v}$. Because $Z=D$, the two residuals are
\begin{align}
    h_{\theta,\rm d}^{(M)}-Y_{\rm d}
      &=(am_Z+w_{\rm l}m_D-1)D,\nonumber\\
    h_{\theta,\rm j}^{(M)}-Y_{\rm j}
      &=\sqrt\lambda R+(j_M[u]_+-1)CH.
    \label{eq:actualresiduals}
\end{align}
The squared vector norm adds the two squared coordinates; no cross-coordinate product appears. Inside the joint square, the cross term averages to zero because $\mathbb E[CH]=0$ and masks are independent of the cues. Also, $D^2=(CH)^2=1$. Write $\alpha=\mathbb E j_M$; since $j_M$ is binary, $\mathbb E(j_M[u]_+-1)^2=(1-\alpha)+\alpha([u]_+-1)^2$. Write $\mathcal L(\theta;Q)$ for the general diagnostic objective under the specified input-visibility law, allowing degenerate laws for single-modality and full inputs. Its exact expression is
\begin{align}
    2\mathcal L(\theta;Q)={}&\mathbb E_M(am_Z+w_{\rm l}m_D-1)^2
                     +\lambda R^2\nonumber\\
                    &+(1-\alpha)+\alpha([u]_+-1)^2.
    \label{eq:unifiedloss}
\end{align}
Under the diagnostic data distribution defined above, substituting
the predictor in Eq.~\eqref{eq:sharedhead} into the training objective
in Eq.~\eqref{eq:objective}, with squared-error loss
$\ell(h,Y)=\frac12\|h-Y\|^2$ and content-independent masks,
gives Eq.~\eqref{eq:unifiedloss}.
For the main-text SCAP setting, visual and language cues are retained
independently with probability $q$, so that $\alpha=q^2$.
The joint component of Eq.~\eqref{eq:unifiedloss} then becomes
$\mathcal L_{\mathrm{mask},\rm j}=(1-\alpha)/2+F_\alpha(b,u)$,
recovering Eq.~\eqref{eq:maskenergy}.
The VA and LA objectives in Eq.~\eqref{eq:stagingobjectives}
follow by fixing the corresponding single-modality masks,
which gives $\alpha=0$; complete inputs give $\alpha=1$.

\section{Flow-Matching Realization of the Diagnostic Model}
\label{app:fmbridge}

This section constructs a flow-matching counterpart of the diagnostic
model in Appendix~\ref{app:diagnostic}. We show that its expected
velocity loss is a positive constant multiple of Eq.~\eqref{eq:unifiedloss},
yielding equivalent gradient-flow dynamics up to a rescaling of
optimization time.

\subsection{Flow-Matching Setup and Target Decomposition}

Use $A=(D,CH)$, standard normal noise $\epsilon\sim\mathcal N(0,I_2)$, and time $t\sim\rho$ on $[0,1)$, with $X_t=(1-t)\epsilon+tA$ and velocity target $U=A-\epsilon$. Noise, time, semantic cues, and masks are independent. The action-side inputs $S,X_t,t$ remain unmasked, with $S$ constant as in the diagnostic.

For either coordinate $r\in\{\rm d,\rm j\}$, the Gaussian likelihoods with equal action priors give
\begin{equation}
    \mu(x,t):=\mathbb E[A_r\mid X_{t,r}=x,t]
      =\tanh\!\left(\frac{tx}{(1-t)^2}\right).
    \label{eq:posterior}
\end{equation}
Since $U_r=(A_r-X_{t,r})/(1-t)$, define the baseline velocity and residual by
\begin{align}
    v_0(x,t)&=\frac{\mu(x,t)-x}{1-t},\qquad
    g(a;x,t)=\frac{a-\mu(x,t)}{1-t},\nonumber\\
    U_r&=v_0(X_{t,r},t)+g(A_r;X_{t,r},t).
    \label{eq:fmresidual}
\end{align}
Thus $v_0(X_{t,r},t)=\mathbb E[U_r\mid X_{t,r},t]$ and $\mathbb E[g(A_r;X_{t,r},t)\mid X_{t,r},t]=0$. Write $v_{0,r}=v_0(X_{t,r},t)$, $g_r(a)=g(a;X_{t,r},t)$, and $\sigma^2=\mathbb E[g_r(A_r)^2]$, common to both coordinates. Then
\begin{equation}
    \mathbb E[g_r(A_r)^2\mid t]
      =\frac{\mathbb E\operatorname{Var}(A_r\mid X_{t,r},t)}{(1-t)^2}.
    \label{eq:residualvariance}
\end{equation}
For $t\leq1/2$, the numerator is at most one; for $t>1/2$, the estimator $X_{t,r}/t$ bounds it by $(1-t)^2/t^2$. Both yield $\mathbb E[g_r(A_r)^2\mid t]\leq4$. The posterior variance is positive for every $t<1$, hence $0<\sigma^2\leq4$.

\subsection{Velocity Model Construction and Loss Equivalence}

Use the diagnostic parameters, initialization, constants, and masks, with $R=b+u+c$. Using the fixed distribution-dependent features $v_{0,r},g_r$ and constant $\sigma$, define
\begin{align}
    v_{\theta,\rm d}^{(M)}
      &=v_{0,\rm d}+\kappa w_{\rm v}m_Zg_{\rm d}(Z)
                     +w_{\rm l}m_Dg_{\rm d}(D),\nonumber\\
    v_{\theta,\rm j}^{(M)}
      &=v_{0,\rm j}+\sigma\sqrt\lambda R
                     +j_M[u]_+g_{\rm j}(CH).
    \label{eq:fmhead}
\end{align}
Features are evaluated only from retained cues and the unmasked action-side inputs; $g_{\rm j}(CH)$ requires both $C$ and $H$. Subtracting Eq.~\eqref{eq:fmresidual} yields
\begin{align}
    v_{\theta,\rm d}^{(M)}-U_{\rm d}
      &=(\kappa w_{\rm v}m_Z+w_{\rm l}m_D-1)g_{\rm d}(D),\nonumber\\
    v_{\theta,\rm j}^{(M)}-U_{\rm j}
      &=\sigma\sqrt\lambda R+(j_M[u]_+-1)g_{\rm j}(CH).
    \label{eq:fmerrors}
\end{align}
Using mask independence, $\mathbb E[g_r(A_r)]=0$, and $\mathbb E[g_r(A_r)^2]=\sigma^2$, the squared residuals give
\begin{equation}
    \mathcal L^{\rm FM}(\theta;Q)
       :=\tfrac12\mathbb E\|v_\theta^{(M)}-U\|^2
        =\sigma^2\mathcal L(\theta;Q).
    \label{eq:fmcorrespondence}
\end{equation}
Consequently, $\nabla_\theta\mathcal L^{\rm FM}=\sigma^2\nabla_\theta\mathcal L$, and gradient-flow trajectories agree under $T=\sigma^2T_{\rm FM}$. The closing time is divided by $\sigma^2$, the persistent joint-loss floor becomes $\sigma^2/2$, and the protection condition $\lambda(1+c)^2<\alpha$ is unchanged, with $\alpha=q^2$ for SCAP and $\alpha=1$ for full inputs.

After closure, the joint velocity is $v_0(x_{\rm j},t)+\sigma\sqrt\lambda R$, independent of $C,H$. A coordinate-wise sampler with context-independent noise and solver randomness therefore produces the same joint-coordinate distribution for every $(C,H)$. It cannot match both required point masses $\delta_{+1}$ and $\delta_{-1}$; context-independent thresholding gives average sign error $1/2$.

\section{Damage to Joint Vision--Language Modeling}
\label{app:closure}

This appendix provides the detailed derivations for
Sec.~\ref{sec:irreversible}, covering joint-pathway closure during
single-modality pretraining and its persistence under subsequent
full-input training.

\subsection{Joint-Pathway Closure under Single-Modality Pretraining}

% 逻辑：由单模态 alpha 为零求两个真实梯度，再相加相减获得完整积分解，不跳到死门结论。
We now solve the single-modality dynamics from $(b,u)=(0,1)$. Either VA or LA has $j_M=0$. With $R=b+u+c$, independence of the fair signs gives $\mathbb E[CH]=0$ and $(CH)^2=1$, hence
\begin{equation}
\begin{aligned}
    \mathcal L_{\rm single,j}
      &=\tfrac12\mathbb E[(\sqrt\lambda R-CH)^2]\\
      &=\tfrac\lambda2 R^2-\sqrt\lambda R\,\mathbb E[CH]\\
      &\quad+\tfrac12\mathbb E[(CH)^2]\\
      &=\tfrac12(\lambda R^2+1).
\end{aligned}
\label{eq:singlejointloss}
\end{equation}
The direct-coordinate block does not involve $b,u$. Differentiate with respect to these two independent parameters before using any relation along the trajectory:
\begin{equation}
\begin{aligned}
    \partial_b\mathcal L&=\lambda R\,\partial_bR=\lambda R,\\
    \partial_u\mathcal L&=\lambda R\,\partial_uR=\lambda R.
\end{aligned}
\label{eq:singlejointgradients}
\end{equation}
Equation~\eqref{eq:singlejointflow} follows from negative-gradient updates. Adding and subtracting them gives
\begin{align}
    \dot R&=\dot b+\dot u=-2\lambda R,& R(0)&=1+c,\nonumber\\
    \tfrac{\mathrm d}{\mathrm dT}(u-b)&=0,&u(0)-b(0)&=1.
    \label{eq:stageodes}
\end{align}
The first equation can be integrated explicitly:
\begin{equation}
\begin{aligned}
    \tfrac{\mathrm d}{\mathrm dT}\bigl(e^{2\lambda T}R(T)\bigr)&=0,\\
    e^{2\lambda T}R(T)&=R(0)=1+c,\\
    R(T)&=(1+c)e^{-2\lambda T}.
\end{aligned}
\label{eq:singleresidualsolution}
\end{equation}
The second supplies the parameter invariant:
\begin{equation}
\begin{aligned}
    \tfrac{\mathrm d}{\mathrm dT}(u-b)&=\dot u-\dot b=0,\\
    u(T)-b(T)&=u(0)-b(0)=1.
\end{aligned}
\label{eq:singleparameterinvariant}
\end{equation}
Thus $b=u-1$ and $R=2u+c-1$, yielding the gate trajectory in Eq.~\eqref{eq:stageclosure}. These identities hold along the solution and are not constraints imposed during differentiation. Setting $u(T)=0$ gives $T_{\rm close}=\log((1+c)/(c-1))/(2\lambda)$. Since $c>1$, closure occurs in finite time and $u(T)<0$ for every $T>T_{\rm close}$. The missing joint signal does not oppose the offset-fitting updates that close the initially active gate.

\subsection{Persistent Closure under Full-Input Training}

% 逻辑：即使联合项导数为零，共享参数仍可被偏移更新；需证明该剩余更新只会继续把门推负。
The purpose of this step is to rule out recovery for the entire full-input continuation, not merely to observe a negative gate at the switch. Restore full inputs at $T_{\rm p}>T_{\rm close}$. In the region $u<0$, the gate and its derivative vanish, but the offset derivatives remain. Therefore $\dot R=-2\lambda R$ and $\dot u=-\lambda R$. Their candidate continuation is
\begin{align}
    R(T_{\rm p}+T_{\rm a})
      &=R(T_{\rm p})e^{-2\lambda T_{\rm a}}>0,\nonumber\\
    u(T_{\rm p}+T_{\rm a})
      &=u(T_{\rm p})-\tfrac{R(T_{\rm p})}{2}
                      (1-e^{-2\lambda T_{\rm a}})<0.
    \label{eq:trappedtrajectory}
\end{align}
The second line verifies the assumed region for every $T_{\rm a}\geq0$, so the solution never reaches the ReLU boundary. Its limits are $R\to0$ and $u\to(1-c)/2<0$. Substituting into Eq.~\eqref{eq:unifiedloss} with full inputs gives $\mathcal L_{\rm full,j}=(1+\lambda R^2)/2\to1/2$. The gradients already use the complete population distribution, not a finite data-coverage approximation.

% 逻辑：明确同一模型有零损失解，且失败确实是对不同联合上下文无响应而非仅一个抽象损失数字。
At $u=1$, $b=-c-1$, and $\kappa w_{\rm v}+w_{\rm l}=1$, both residuals in Eq.~\eqref{eq:actualresiduals} are identically zero under full inputs. Thus the failure is not a capacity lower bound. The data require $p(A_{\rm j}\mid C,H)=\delta_{CH}$, where $\delta_a$ is a point mass at $a$. After closure, the prediction is the constant $\sqrt\lambda R$, independent of $C,H$. It cannot realize both conditional outputs $+1$ and $-1$; its limiting conditional prediction is zero. The FM realization in Appendix~\ref{app:fmbridge} establishes the analogous context-blindness statement for a generative sampler, separately from this deterministic regression example.

% 逻辑：不可逆针对给定更新规则，不是数据从物理意义上丢失，也不是初始就死的门。
Paired data still reveal the joint relation. The obstruction is the inherited parameter state under the unchanged optimizer, not irreversible destruction of dataset information. A reset, a different shared architecture, or a nonzero negative-side activation derivative can alter the result. An independent gate that starts dead would not show damage caused by pretraining; here $u(0)=1$ and first-stage updates cause closure explicitly.

\section{Preserving Joint Modeling with SCAP}
\label{app:prevention}

\subsection{Joint Loss under SCAP Masking}

For completeness, expand the joint-coordinate expectation before separating the parameter-independent constant. We allow the more general independent probabilities $q_{\rm v},q_{\rm l}\in(0,1)$ here; the main-text setting is recovered by setting both to $q$. Keep the target $CH$ unchanged. The residual is
\begin{equation}
\begin{aligned}
    h_{\theta,\rm j}^{(M)}-CH
      &=\sqrt\lambda R+(j_M[u]_+-1)CH.
\end{aligned}
\label{eq:maskjointresidual}
\end{equation}
At a fixed mask, independence gives $\mathbb E[CH\mid M]=0$ and $\mathbb E[(CH)^2\mid M]=1$, so
\begin{equation}
\begin{aligned}
    &\mathbb E[(h_{\theta,\rm j}^{(M)}-CH)^2\mid M]\\
    &\quad=\lambda R^2
       +2\sqrt\lambda R(j_M[u]_+-1)\mathbb E[CH\mid M]\\
    &\qquad+(j_M[u]_+-1)^2\mathbb E[(CH)^2\mid M]\\
    &\quad=\lambda R^2+(j_M[u]_+-1)^2.
\end{aligned}
\label{eq:maskconditionalsquare}
\end{equation}
Since $j_M$ is binary with $\alpha=\Pr(j_M=1)=q_{\rm v}q_{\rm l}$,
\begin{equation}
\begin{aligned}
    \mathbb E_M[(j_M[u]_+-1)^2]
      &=(1-\alpha)\cdot1+\alpha([u]_+-1)^2,\\
    \mathcal L_{\mathrm{mask},\rm j}
      &=\tfrac12[\lambda R^2+1-\alpha+\alpha([u]_+-1)^2].
\end{aligned}
\label{eq:maskjointaverage}
\end{equation}
Separating the constant yields Eq.~\eqref{eq:maskenergy}. Substituting $\alpha$ for $j_M$ inside the square would lose the mask-variance contribution. In particular, masks missing joint cues still contribute the offset loss as well as the constant $(1-\alpha)/2$.

\subsection{Joint-Pathway Preservation under SCAP Pretraining}

% 逻辑：同一目标去掉的仅为常数，缺少联合线索的 mask 仍贡献偏移训练，防止选择性取样。
We seek a sufficient condition preventing the first boundary hit $u=0$. Use the main-text independent retention with $\alpha=q^2$; the proof also applies to the generalization $\alpha=q_{\rm v}q_{\rm l}$. Equation~\eqref{eq:unifiedloss} gives $\mathcal L_{\mathrm{mask},\rm j}=(1-\alpha)/2+F_\alpha$. Removing the constant does not discard unfavorable masks: every mask still contributes the offset term $\lambda R^2/2$. While $u>0$,
\begin{align}
    \dot b&=-\lambda R,\nonumber\\
    \dot u&=-\lambda R+\alpha(1-u).
    \label{eq:protectedodes}
\end{align}
The second term supplies a restoring signal when offset fitting lowers $u$ below one. A positive restoring signal by itself need not dominate the downward term; the barrier condition below provides a sufficient quantitative guarantee.

% 逻辑：先计算初始和边界，再利用梯度流连续且不升高目标值排除第一次触界。
Assume $\lambda(1+c)^2<\alpha\leq1$. At $(b,u)=(0,1)$, $F_\alpha=\lambda(1+c)^2/2<\alpha/2$. At the closing boundary,
\begin{equation}
\begin{aligned}
    F_\alpha(b,0)
      &=\tfrac12[\lambda(b+0+c)^2+\alpha(1-0)^2]\\
      &=\tfrac\lambda2(b+c)^2+\tfrac\alpha2\\
      &\geq\tfrac\alpha2,\qquad\forall b\in\mathbb R.
\end{aligned}
\label{eq:closingboundarybound}
\end{equation}
The lower bound uses $\lambda>0$ and $(b+c)^2\geq0$, with equality at $b=-c$; the strict protection condition instead places the initialization below this bound. Before any hypothetical first boundary hit, $\dot F_\alpha=-\|\nabla F_\alpha\|^2\leq0$. Continuity would make the loss at that hit both smaller than and at least $\alpha/2$, a contradiction. Thus $u(T)>0$ for all finite $T$.

% 逻辑：保护通路不等于拟合完成，因此继续证明正区域目标严格凸且最优点恰好恢复联合关系。
Avoiding closure does not by itself prove successful fitting. To establish convergence to an exact joint predictor, compute the Hessian in the active region:
\begin{equation}
    G_\alpha=\begin{pmatrix}\lambda&\lambda\\\lambda&\lambda+\alpha\end{pmatrix},
    \qquad\det G_\alpha=\lambda\alpha>0.
    \label{eq:gatehessian}
\end{equation}
Its leading diagonal entry is positive, so the matrix is positive definite. Setting both gradients to zero gives $R=0$ and $u=1$, hence $b=-c-1$. This strictly convex quadratic has bounded sublevel sets and exponentially convergent gradient flow. The trajectory therefore converges to the unique exact joint-relation solution rather than merely avoiding closure.

\subsection{Joint-Pathway Preservation under Full-Input Training}

% 逻辑：切换目标后旧能量不再直接适用，显式比较 F1 和 Falpha 再对新边界重复论证。
We must transfer the boundary argument to the new full-input objective; decreasing the old objective alone would not justify protection after the switch. At any finite switch time $T_{\rm p}$, $F_\alpha(T_{\rm p})\leq F_\alpha(0,1)<\alpha/2$. For $0<\alpha\leq1$,
\begin{align}
    F_1(b,u)&=\tfrac12[\lambda R^2+(1-[u]_+)^2]\nonumber\\
      &\leq\tfrac1{2\alpha}[\lambda R^2+\alpha(1-[u]_+)^2]
       =F_\alpha(b,u)/\alpha.
    \label{eq:switchbarrier}
\end{align}
Hence $F_1(T_{\rm p})<1/2$, below every point on the boundary $u=0$ for the new objective. The full-input trajectory stays active and converges to $(b,u)=(-c-1,1)$, with zero joint loss. The same argument with $\alpha=1$ protects full-input training from the original initialization. Masking is not uniquely necessary for avoiding the trap.

% 逻辑：给两个定理同时成立的非空例子，再强调 mask 不能事后复活以及纯单模态交替仍会关闭。
The following numerical example only demonstrates that the sufficient condition is feasible; it is not an experiment or an additional proof assumption. For $c=2$, $\lambda=1/64$, and $q=1/2$,
\begin{equation}
\begin{aligned}
    \lambda(1+c)^2&=\tfrac1{64}(1+2)^2=\tfrac9{64}\\
      &<\tfrac14=q^2=\alpha.
\end{aligned}
\label{eq:protectionexample}
\end{equation}
A common first-stage budget $T_{\rm p}=60>32\log3=T_{\rm close}$ closes the gate under either single-modality route while the masked route remains protected. Alternating only VA and LA also has $j_M=0$ and the same closing dynamics. Once $u<0$, adding masks does not restore the missing joint derivative; the continuation in Eq.~\eqref{eq:trappedtrajectory} still applies. The result concerns prevention, not post-closure repair. Preserving pretrained generalization during video adaptation is a related but distinct objective~\cite{zhu2024mote,zhu2026taco}; our result concerns an explicitly modeled joint pathway.

\section{Reducing Visual Shortcuts with SCAP}
\label{app:direct}

Related causal work also targets spurious correlations in category-level pose estimation~\cite{lin2025cleanpose}; here we isolate training-route effects in a controlled cue model. Both routes start at $w_{\rm v}=w_{\rm l}=0$. Define $a=\kappa w_{\rm v}$, $d=w_{\rm l}$, $g=\kappa^2$, and $\beta=g+1$. Only input visibility changes; a finite comparison uses total duration $T_{\rm p}+T_{\rm a}$ for both routes. In this appendix $\mathcal L(\theta;Q)$ denotes the general objective for a specified mask law, whereas route-specific losses retain the subscripts $\rm full$ and $\rm mask$.
The three core parts assume $g>1$ and independent common retention $q\in(0,1)$.

\subsection{Visual Dominance under Full-Input Training}
We first recover the time-dependent full-input baseline used in the route comparison. Because $Z=D$ and $D^2=1$, the loss and its original-parameter derivatives are
\begin{equation}
\begin{aligned}
    \mathcal L_{\rm full,d}
      &=\tfrac12\mathbb E[(\kappa w_{\rm v}Z+w_{\rm l}D-D)^2]\\
      &=\tfrac12(a+d-1)^2,\\
    \partial_{w_{\rm v}}\mathcal L_{\rm full,d}&=\kappa(a+d-1),\\
    \partial_{w_{\rm l}}\mathcal L_{\rm full,d}&=a+d-1.
\end{aligned}
\label{eq:fulldirectgradients}
\end{equation}
Since $\dot a=\kappa\dot w_{\rm v}$, the dynamics are Eq.~\eqref{eq:fullrouteodes}; dropping this extra factor would change the optimizer. The invariant $a-gd=0$ and the scalar equation $\dot s=\beta(1-s)$ for $s=a+d$, with $s(0)=0$, give $s=1-\exp(-\beta T)$. Solving for both coefficients yields Eq.~\eqref{eq:fullroutesolution}. This baseline uses neither a visual-only reference point nor a preceding single-modality phase.
Thus $a_{\rm F}^{\infty}=g/\beta>d_{\rm F}^{\infty}=1/\beta$ when $g>1$: the visual cue dominates even though the full-input training error vanishes.

\subsection{Language-Dominant Solutions with SCAP}

Let $(a_{\rm p},d_{\rm p})$ be the actual endpoint after a finite masked pretraining duration $T_{\rm p}>0$, without assuming convergence. Continuing with complete visual and language inputs gives the full-stage limits
\begin{align*}
    a_{\rm MF}^{\infty}&=\tfrac{g+a_{\rm p}-gd_{\rm p}}\beta,\\
    d_{\rm MF}^{\infty}&=\tfrac{1+gd_{\rm p}-a_{\rm p}}\beta.
\end{align*}
These are Eq.~\eqref{eq:fullendpointlimits}. Comparing them with the full-input baseline gives
\begin{align*}
    d_{\rm MF}^{\infty}-d_{\rm F}^{\infty}
      &=\tfrac{gd_{\rm p}-a_{\rm p}}\beta,\\
    a_{\rm MF}^{\infty}-a_{\rm F}^{\infty}
      &=-\tfrac{gd_{\rm p}-a_{\rm p}}\beta.
\end{align*}
The sign of this common numerator determines whether MF increases language reliance and reduces visual reliance. We therefore define $\delta(T)=gd(T)-a(T)$ and $\delta_{\rm p}=gd_{\rm p}-a_{\rm p}$, as in Eq.~\eqref{eq:deltadefinition}.

To justify the full-stage limits, write $r=1-a-d$. Full-input updates give $\dot r=-\beta r$, hence $a+d\to1$, while the inherited difference remains constant:

\begin{equation}
    \left.\dot\delta\right|_{\rm full}
      =g\dot d-\dot a=gr-gr=0.
    \label{eq:fullinvariant}
\end{equation}

Solving $a+d=1$ and $gd-a=\delta_{\rm p}$ yields the limits above. It remains to prove $\delta_{\rm p}>0$ along the actual masked trajectory. Expand the masked direct-coordinate error before taking expectations:

\begin{equation}
\begin{aligned}
    &(am_Z+dm_D-1)^2\\
    &\quad=a^2m_Z^2+d^2m_D^2+2ad m_Zm_D\\
    &\qquad-2am_Z-2dm_D+1.
\end{aligned}
\label{eq:maskdirectsquare}
\end{equation}

Using $m_Z^2=m_Z$, $m_D^2=m_D$, $\mathbb E m_Z=\mathbb E m_D=q$, and $\mathbb E m_Zm_D=q^2$ gives Eqs.~\eqref{eq:maskdirectloss} and~\eqref{eq:maskdirectodes}. With $x=(a,d)^\top$ and $f=(gq,q)^\top$, the matrix and equilibrium are
\begin{equation}
\begin{aligned}
    K&=\begin{pmatrix}gq&gq^2\\q^2&q\end{pmatrix},&
    x^*&=\tfrac1{1+q}\begin{pmatrix}1\\1\end{pmatrix}.
\end{aligned}
\label{eq:maskdirectmatrix}
\end{equation}
Since $Kx^*=f$, the integrating-factor identity $\frac{\mathrm d}{\mathrm dT}[\exp(KT)(x-x^*)]=0$, with $x(0)=0$, yields
\begin{align}
    x_{\rm mask}(T)&=(I-\exp(-KT))x^*,\label{eq:masktimecourse}\\
    \begin{pmatrix}a_{\rm p}\\d_{\rm p}\end{pmatrix}
       &=x_{\rm mask}(T_{\rm p})=(I-\exp(-KT_{\rm p}))x^*.
    \label{eq:maskdirecttrajectory}
\end{align}
Here $T$ is optimization time, not an action-sequence index. In particular, the endpoint passed to full-input training is the actual value at finite $T_{\rm p}$ rather than $x^*$.

Under symmetric retention,
\begin{equation}
\begin{aligned}
    \left.\dot\delta\right|_{\rm mask}
      &=gq[(1-d-qa)-(1-a-qd)]\\
      &=gq(1-q)(a-d),\qquad\delta(0)=0.
\end{aligned}
\label{eq:deltamaskflow}
\end{equation}
Therefore it suffices to show that $z=a-d$ is positive. Subtracting the two mask updates and differentiating gives
\begin{equation}
\begin{aligned}
    \dot z&=q(g-1)-q(g-q)a-q(gq-1)d,\\
    \ddot z&=-q(g-q)\dot a-q(gq-1)\dot d\\
      &=-q(g+1)\dot z-gq^2(1-q^2)z.
\end{aligned}
\label{eq:maskdifferenceexpansion}
\end{equation}
Substitution and collection of terms yields the scalar initial-value problem
\begin{equation}
\begin{aligned}
    \ddot z+q(g+1)\dot z+gq^2(1-q^2)z&=0,\\
    z(0)=0,\qquad\dot z(0)&=q(g-1).
\end{aligned}
\label{eq:maskdifferenceode}
\end{equation}
Its two positive decay rates are
\begin{equation}
\begin{aligned}
    \rho_\pm&=\tfrac q2\left[g+1\pm\sqrt{(g-1)^2+4gq^2}\right],\\
    &\hspace{2em}0<\rho_-<\rho_+.
\end{aligned}
\label{eq:maskdecayrates}
\end{equation}
The general solution $z=A_-\exp(-\rho_-T)+A_+\exp(-\rho_+T)$ satisfies $A_+=-A_-$ and $(\rho_+-\rho_-)A_-=q(g-1)$. Hence
\begin{align}
    a(T)-d(T)&=\frac{q(g-1)}{\rho_+-\rho_-}
            [\exp(-\rho_-T)-\exp(-\rho_+T)]>0,\nonumber\\
    \delta_{\rm p}
      &=gq(1-q)\int_0^{T_{\rm p}}(a(T)-d(T))\,\mathrm dT>0.
    \label{eq:maskadvantage}
\end{align}
Both the coefficient and exponential difference are positive for $T>0$. Integrating with $\delta(0)=0$ gives the following closed expression for every finite $T_{\rm p}>0$:
\begin{align}
    \delta_{\rm p}
      &=\frac{gq^2(1-q)(g-1)}{\rho_+-\rho_-}\nonumber\\
      &\quad\times\left[\frac{1-\exp(-\rho_-T_{\rm p})}{\rho_-}
                   -\frac{1-\exp(-\rho_+T_{\rm p})}{\rho_+}\right]>0.
    \label{eq:advantageclosed}
\end{align}
Since $a,d\to1/(1+q)$, the strictly increasing $\delta(T)$ tends to $(g-1)/(1+q)$. Hence $0<\delta_{\rm p}<(g-1)/(1+q)<g$ for every finite positive mask duration. At $g=1$, symmetry instead gives $a=d$ and $\delta=0$: there is no strict limiting advantage from this symmetric schedule.

The positive sign established above therefore gives, for every fixed finite $T_{\rm p}>0$,
\begin{align}
    a_{\rm MF}^{\infty}&=\tfrac{g-\delta_{\rm p}}\beta
                           <\tfrac g\beta=a_{\rm F}^{\infty},\nonumber\\
    d_{\rm MF}^{\infty}&=\tfrac{1+\delta_{\rm p}}\beta
                           >\tfrac1\beta=d_{\rm F}^{\infty}.
    \label{eq:routelimits}
\end{align}

We next determine when relative language improvement becomes language dominance. Taking the separate limit $T_{\rm p}\to\infty$ of the already derived full-stage outcome gives, under symmetric retention,
\begin{align}
    a_{\rm MF}^{\infty,\infty}&=\frac{gq+1}{(g+1)(1+q)},\nonumber\\
    d_{\rm MF}^{\infty,\infty}&=\frac{g+q}{(g+1)(1+q)}.
    \label{eq:longmaskroutes}
\end{align}
For a finite mask stage, language dominates vision at the full-stage limit exactly when $\delta_{\rm p}>(g-1)/2$. Such stages exist because $(g-1)/(1+q)>(g-1)/2$. Nevertheless the visual coefficient in Eq.~\eqref{eq:longmaskroutes} is positive, so neither this limit nor the finite-stage result eliminates the shortcut completely. 

\subsection{Lower Loss under Visual-Cue Flips}

To assess reliance on the visual shortcut, we flip $Z=D$ to $Z=-D$ while keeping the language instruction, target, and other cues unchanged. The shortcut cue now conflicts with the instruction. The bound $0<\delta_{\rm p}<(g-1)/(1+q)<g$ proved above and $a_{\rm MF}^{\infty}=(g-\delta_{\rm p})/\beta$ imply $0<a_{\rm MF}^{\infty}<a_{\rm F}^{\infty}$.

Under this intervention, the direct-coordinate loss is $\mathcal E_{\rm flip}(a,d)=\frac12(1+a-d)^2$ by Eq.~\eqref{eq:flipmetric}. Both full-stage limits satisfy $a+d=1$ on the training distribution, so substituting $d=1-a$ gives $\mathcal E_{\rm flip}=2a^2$. Therefore
\begin{align*}
    \mathcal E_{\rm flip,MF}^{\infty}
      &=2(a_{\rm MF}^{\infty})^2\\
      &<2(a_{\rm F}^{\infty})^2
       =\mathcal E_{\rm flip,F}^{\infty}.
\end{align*}
This proves Eq.~\eqref{eq:shortcutcomparison} for every finite $T_{\rm p}>0$: after full-input convergence, MF follows the unchanged instruction more robustly under this visual-cue perturbation.

\section{Token-Masking Implementation}
\label{app:token-masking}

\paragraph{Sampling.}
For each sample, we pool the expanded image-content positions and instruction tokens into $N$ eligible positions and uniformly select $K=\lfloor Nr+0.5\rfloor$ for embedding zeroing without replacement, with $r=0.3$ by default. This is fixed-count sampling over both modalities, not independent Bernoulli sampling or a separate quota per modality. System-prompt tokens, role and vision delimiters, and padding are excluded. All instruction-content embeddings may be zeroed; no minimum-content constraint is imposed. At $r=1$, all eligible content embeddings are zeroed, while protected-token embeddings and the separately supplied state remain available.

\paragraph{Encoding and conditioning.}
All valid images first pass through the vision encoder. Selected multimodal token embeddings are then zeroed in place before the language backbone, retaining sequence positions and original multidimensional RoPE coordinates. This is content-embedding masking, not token deletion, sequence compaction, or replacement with a learned mask embedding. Sequence-validity masks accompany the inputs through the backbone and action decoder to identify padding. The expert reads the full-attention-layer K/V from this masked-input forward pass. Zero input embeddings need not yield zero hidden states or K/V, since subsequent contextual processing can propagate retained information to masked positions. No retention-dependent rescaling or additional embedding-dimension dropout is applied. Masking does not reduce vision-encoder computation, and unmasked visual features may encode information from other image regions.

\paragraph{Validation and theoretical scope.}
Stage-1 validation retains masking with a separate fixed-seed random stream; evaluation mode alone does not disable this preprocessing. Stage-2 training, downstream benchmark evaluation, and inference use complete inputs. With $R=N-K$ eligible tokens left unmasked, two distinct eligible tokens are jointly unmasked with probability $R(R-1)/[N(N-1)]$ for $N>1$, rather than $(R/N)^2$. These token-level probabilities are not the diagnostic's semantic-cue probabilities: the independent-cue analysis characterizes the mechanism, not an exact guarantee for the implemented VLA.

\section{Training Data and Preprocessing}
\label{app:training-data}

\subsection{Source Composition and Accounting}

The audited collection combines ABC-130k~\cite{allshire2026abc}, AgiBotWorld~\cite{bu2025agibotworld,agibot2026worlddata}, DreamZero-DROID~\cite{gear2026dreamzerodroid} (a DROID-derived collection~\cite{khazatsky2024droid}), Galaxea Open-World~\cite{jiang2025galaxea}, GenieSim~\cite{yin2026geniesim}, RoboCOIN~\cite{wu2025robocoin}, RoboMIND~\cite{wu2024robomind,hou2025robomind2}, and InternData-A1~\cite{tian2025interndata}, a public synthetic dataset. It also includes the RoboTwin and RoboCasa benchmark collections. The audit reports 9,843.16 retained hours across 3,579 data repositories; repositories are storage units, not distinct tasks or embodiments. Table~\ref{tab:v2-source-composition} separates source-level duration from the approximately 1.2M trajectory budget in Sec.~\ref{sec:training-data}. Hours do not determine an episode count or the number of sampled action chunks.

\begin{table}[t]
    \centering
    \small
    \begin{tabular}{@{}lrr@{}}
        \hline
        Source & Original (h) & Retained (h) \\
        \hline
        ABC-130k~\cite{allshire2026abc} & 4,126.29 & 1,642.30 \\
        AgiBotWorld-Beta~\cite{bu2025agibotworld} & 2,344.34 & 2,286.54 \\
        AgiBotWorld 2026~\cite{agibot2026worlddata} & 440.70 & 390.66 \\
        DreamZero-DROID~\cite{gear2026dreamzerodroid} & 271.41 & 263.80 \\
        Galaxea Open-World~\cite{jiang2025galaxea} & 454.61 & 423.37 \\
        GenieSim~\cite{yin2026geniesim} & 230.51 & 211.38 \\
        RoboCOIN~\cite{wu2025robocoin} & 677.17 & 653.23 \\
        RoboMIND 1.0~\cite{wu2024robomind} & 192.53 & 36.91 \\
        RoboMIND 2.0~\cite{hou2025robomind2} & 1,166.89 & 262.37 \\
        InternData-A1~\cite{tian2025interndata} & 3,721.71 & 3,559.02 \\
        \hline
        RoboCasa-GR1~\cite{robocasaGR1Tabletop} & 80.84 & 80.84 \\
        RoboTwin Clean~\cite{chen2025robotwin2} & 2.98 & 2.98 \\
        RoboTwin Randomized~\cite{chen2025robotwin2} & 29.77 & 29.77 \\
        LeRobot OXE~\cite{oxe2023rtx,cadene2024lerobot} & 935.86 & 0.00 \\
        \hline
        Audit total & 14,675.59 & 9,843.16 \\
        \hline
    \end{tabular}
    \caption{\textbf{Dataset composition before and after preprocessing.} Durations are audit statistics, not stage-wise sampling weights. Benchmark collections are shown separately and their inclusion here does not establish pretraining use. Independently rounded source rows differ from the audit totals by at most 0.02 h.}
    \label{tab:v2-source-composition}
\end{table}

\subsection{State and Action Conversion}

The shared storage layout contains 14 arm-joint channels, 14 end-effector (EE) pose channels, two gripper channels, 12 dexterous-hand channels, and four torso channels, totaling 46 dimensions. Arm and hand blocks are ordered left before right. Each arm has at most seven joint channels; six-joint arms and the absent arm of a single-arm embodiment are padded. Each absolute EE pose is represented as $[x,y,z,q_x,q_y,q_z,q_w]$. Euler-angle and rotation-6D sources require source-specific quaternion conversion, normalization, and temporal sign-continuity checks. Where supported, forward kinematics (FK) supplies EE poses from joint targets; native poses are otherwise retained. Padding is not an observed control target, and a shared layout does not imply complete annotations for every channel.

Gripper values are calibrated to $[0,1]$, with zero fully closed and one fully open; source ranges and opening directions are handled separately. Each hand uses little-, ring-, middle-, and index-finger flexion, followed by thumb bending and rotation. Supported mappings retain explicit finger semantics; unresolved mappings are zero-filled rather than assigned guessed meanings. Torso slots follow yaw, pitch, roll, and lift. Source torso fields that cannot be mapped unambiguously are not represented as verified torso controls. The 46-dimensional layout does not include additional head or mobile-base fields. Conversion to a common layout is distinct from action normalization or the loss treatment of unavailable channels, whose exact settings require the training configuration.

\subsection{Quality Checks and Retention Exceptions}

The quality audit assesses six failure types: absent EE states throughout a trajectory; missing one side of a dual-arm EE annotation; excessive disagreement between FK and recorded EE poses; abrupt state/action changes; inconsistent state--action motion trends even after allowing temporal lag; and extreme or non-finite values. A single-arm trajectory only requires its active side. The largest reported exclusions are 2,421.36 h lacking EE data and 2,000.94 h that could not be validated; FK, abrupt-change, trend, and numerical checks account for additional removals. LeRobot OXE contributes no retained data, so its appearance in the earlier conversion inventory is not evidence of use in SCAP pretraining.

These checks do not certify every retained field. The benchmark collections have no reported rejection breakdown, and the RoboCasa collection retains 80.84 h without unified EE annotations. Some RoboMIND checks validate position but not the complete orientation convention.

\subsection{Coordinate Conventions and Reporting Scope}

The preprocessing pipeline preserves native pose frames and tool reference points. Table~\ref{tab:v2-coordinate-frames} quantifies this heterogeneity. For example, AgiBotWorld-Beta uses head-camera poses at the wrist/flange, whereas AgiBotWorld 2026 uses an arm-base frame and an end-link reference with a tool offset. DROID-derived poses use the Franka base and link-8 reference; Galaxea uses a torso frame and arm-link-6 reference. Different RoboMIND and RoboCOIN sources retain distinct base/world frames and tool offsets. Consequently, a common quaternion format does not establish a common wrist or tool-center-point convention. Additional camera-frame and wrist-reference transformations are not part of the collection described here.

\begin{table}[t]
    \centering
    \small
    \begin{tabular}{@{}lrr@{}}
        \hline
        EE frame category & Hours & Share (\%) \\
        \hline
        Base/torso local & 6,902.46 & 70.12 \\
        Head camera & 2,286.54 & 23.23 \\
        World/simulation scene & 290.94 & 2.96 \\
        Unresolved frame/reference & 282.39 & 2.87 \\
        Unified EE unavailable & 80.84 & 0.82 \\
        \hline
        Total & 9,843.16 & 100.00 \\
        \hline
    \end{tabular}
    \caption{\textbf{EE coordinate conventions in the data collection.} Categories describe stored annotations, not globally aligned geometry. Values are rounded independently.}
    \label{tab:v2-coordinate-frames}
\end{table}

\FloatBarrier
\section{Real-Robot Evaluation Details}
\label{app:real-robot-details}

Each of the seven initialization routes is adapted independently to three task families, yielding three task-specific checkpoints per route. The dataset contains 300 demonstrations, and evaluation comprises 20 episodes per task-specific policy, 60 per route, and 420 across all seven routes. Results for all three tasks are reported in Table~\ref{tab:robot-tabletop-results}.

\subsection{Hardware, Observations, and Control}
\label{app:robot-hardware}

\begin{figure}[!t]
    \centering
    \includegraphics[width=\columnwidth]{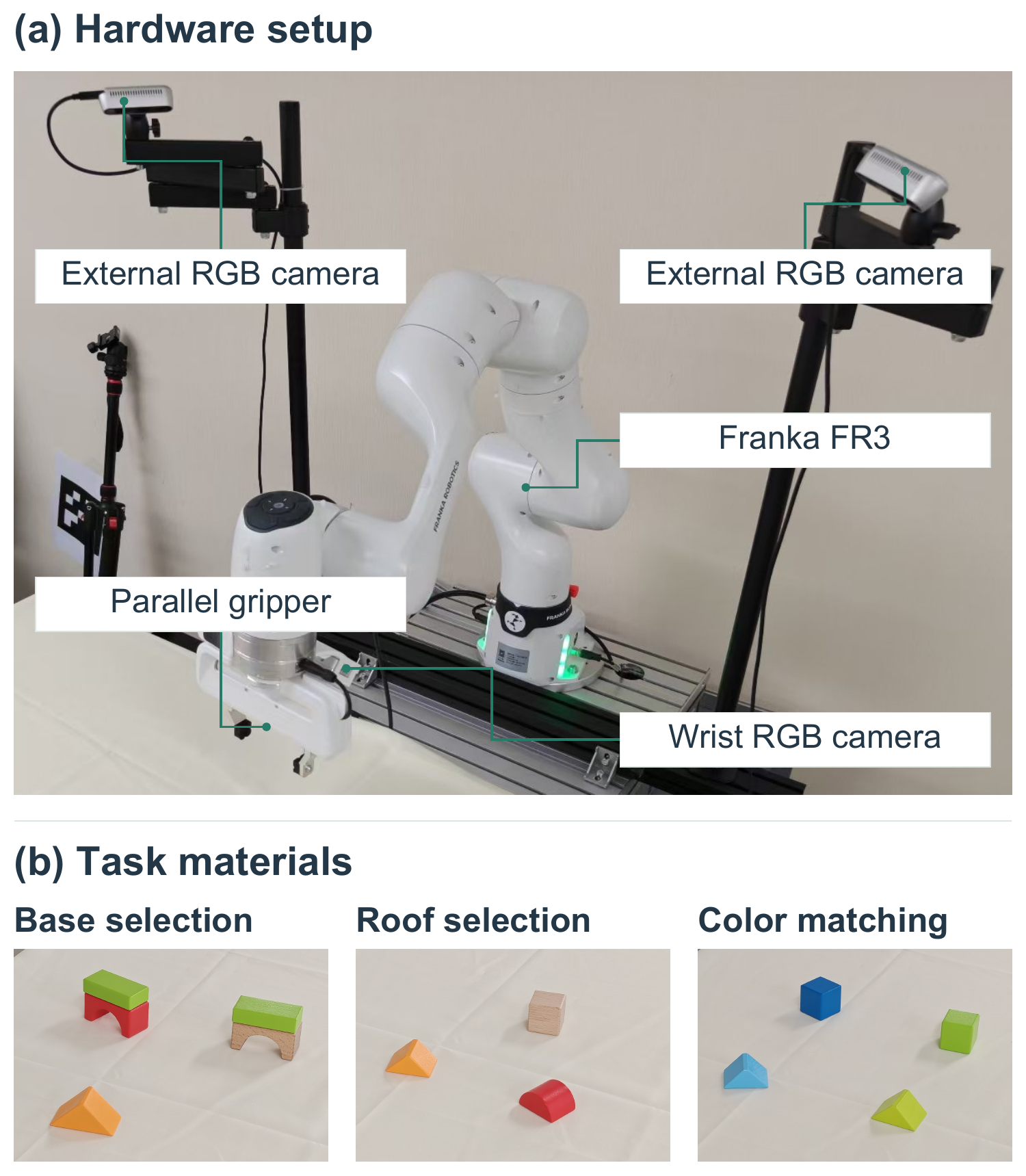}
    \caption{\textbf{Real-robot hardware and task materials.} (a) Franka Research 3, parallel gripper, two fixed external RGB cameras, and wrist RGB camera. (b) Objects used for base selection, roof selection, and color matching. The photographs document the setup and materials.}
    \label{fig:robot-setup-reference}
\end{figure}

Figure~\ref{fig:robot-setup-reference} shows the hardware and task materials. The platform uses a Franka Research 3 (FR3; controller identifier \texttt{fr3v2.1}) and a two-finger parallel gripper. The three RGB views are \texttt{head} (Intel RealSense D435, fixed external main view), \texttt{exterior\_1} (D435i, fixed external side view), and \texttt{wrist} (D405, wrist-mounted close view). Original images are $640\times480$ and are resized directly to $224\times224$ using float32 bilinear interpolation with \texttt{align\_corners=False}, without cropping or padding. State contains seven joint angles in radians and total gripper opening in meters. State is supplied to the action expert. Geometric and memory-augmented object pose models~\cite{lin2024transpose,lin2026mempose} provide complementary perception capabilities, but are not used to construct our policy inputs.

Recording and upper-level joint-target dispatch are configured at 30\,Hz. Policy inference is queue-driven: a new chunk is predicted after the current action queue is exhausted. The upper layer limits each joint-target increment to $\pm0.005$\,rad without explicit linear interpolation; the underlying franky controller uses Ruckig trajectory generation. The current implementation has no fixed overall episode time limit. Safety watchdog timeouts are not treated as episode-duration limits. Singularity-consistent arm planning~\cite{zhong2025singularity} addresses kinematic trajectory constraints beyond our context-conditioning objective; that planner is not used by our controller. Motion planning for zero-shot continuous vision-and-language navigation~\cite{liu2026stamp} addresses a different task setting; our manipulation setup does not include navigation or mapping.
% One training seed is used. Do not describe a multi-seed experiment.
% Gripper model remains unverified. Prefix alignment and chunk32 are the common deployment settings.

\subsection{Demonstrations and Idle-Frame Removal}

Each task contains two instruction conditions with 50 demonstrations each. Base selection refers to the red or natural-wood base; roof selection refers to the orange triangular or red rounded roof. Color matching counts both roof placements as one episode. All three task datasets use the same idle-frame removal rule. A transition is idle only when the maximum absolute change across the seven joint angles is below $10^{-4}$\,rad and the gripper-width change is below $10^{-4}$\,m. Leading, trailing, and internal near-stationary transitions are removed; adjacent retained frames are rechecked iteratively. There is no minimum stationary-duration requirement or fixed waiting-frame window around gripper events.

The procedure retains the starting frames of active transitions and the final active target frame, while selecting all three images and state synchronously. If original retained indices are $r_0,r_1,\ldots$, the reconstructed state is $x[r_k]$ and the action target is $x[r_{k+1}]$. Labels therefore point to the next retained measured state, not necessarily the original next frame. The last observation has no successor and is not assigned a cross-episode action target. Original frame mappings, timestamps, and inter-frame intervals are retained, even though exported video is rearranged at 30 fps. A 32-target chunk consequently has no fixed original-time duration.

Conversion checks preserve the number of gripper opening/closing events and inspect jumps across removed intervals. Additional joint, gripper, and end-effector translation changes relative to the original interval's largest single-step change are bounded by $0.0005$\,rad, $0.0001$\,m, and $0.0005$\,m, respectively. These are splice checks, not a justification for removing contact waits during deployment. Episode lists and held-out test layouts are fixed separately from checkpoint selection.

\subsection{Joint Targets and Task-Specific Normalization}

Action labels are absolute next-retained-observation joint positions and gripper opening, not Cartesian velocities or controller setpoints. The 46D projector interface stores seven joint angles at zero-based coordinates 0--6 and normalized gripper opening at coordinate 28. Other coordinates are zero-filled and invalid; masks exclude them from supervision. Gripper opening is mapped by $\operatorname{clip}((w-w_{\rm closed})/(w_{\rm open}-w_{\rm closed}),0,1)$. State uses $\operatorname{sign}(x)\log(1+|x|)$ inside the projector, with mean-zero/unit-scale state statistics. Valid action coordinates use $(a-\mu)/\max(\sigma,10^{-6})$; a horizon mask excludes padding at episode boundaries.

Gripper endpoints and action statistics are task-specific, rather than pooled across all three tasks. Within each task, all seven model routes share the same endpoints and statistics, computed from that task's training data. Each checkpoint retains the corresponding mapping and normalization for inference. This matches separate task-specific adaptation and does not require cross-task normalization. Shared state/action projection modules can be transferred together with the pretrained expert.

\subsection{Initialization and Frozen-VLM Adaptation}

In Table~\ref{tab:robot-tabletop-results}, each indented row denotes an independent pretraining route built on the same Qwen backbone. Method names follow the simulation comparison.

Real-robot runs use seven initialization routes aligned with the simulation comparison: the base Qwen3.5-9B without action-expert pretraining, Full, LA$\rightarrow$Full, VA$\rightarrow$Full, VL-Mask$\rightarrow$Full, and the frozen-VLM and Ft-VLM SCAP$\rightarrow$Full variants. The Qwen baseline initializes the action expert and state/action projections randomly. All other routes transfer their corresponding pretrained expert and projection weights. Frozen-VLM and Ft-VLM label the pretraining stage; all seven routes freeze the VLM during real-robot SFT. Since the two SCAP initializations may differ in pretraining data or budget, their comparison alone does not isolate the effect of updating the VLM. Each route's initialization is copied independently for the three task-specific adaptations, retaining its VLM weights while optimizing the expert and state/action projections. Pretrained vision--language knowledge has also been transferred to category-level pose estimation~\cite{lin2024clipose}; the transfer studied here targets action prediction.

The expert is a 32-layer hybrid-attention MoT with hidden width 1024 and MLP width 4096. Adaptation predicts 32 actions, with no action-position embedding. Each task-specific policy is trained for 10,000 updates on eight GPUs with per-GPU batch size 8 and no gradient accumulation (global batch 64). AdamW uses learning rate $5\times10^{-5}$, weight decay 0.01, gradient clipping 1.0, and a 500-update warmup followed by cosine decay. VLM weights use BF16; training uses DeepSpeed ZeRO-1 with expert and VLM gradient checkpointing disabled. Evaluation always uses the final 10,000-update checkpoint.
% Record exact source IDs and full precision configuration with experiment artifacts.

\subsection{Paired Layouts and Counterbalancing}

Each route uses the same 10 held-out layouts per task, with two instructions evaluated separately after resetting the scene, policy history, and action cache. Instruction order is randomized. How SCAP interacts with explicit motion-aware representations~\cite{zhu2023fima} and persistent multimodal memory~\cite{chen2025telemem} in longer-horizon tasks remains an open question.

For base selection, the red and natural-wood bases exchange positions across layouts; the instruction names the base color, not a spatial location. For roof selection, the triangular and rounded roofs exchange candidate positions across layouts. For color matching, roof positions and body positions vary independently, so that a fixed same-color or different-color instruction requires different spatial pairings across layouts.
\subsection{Instruction-Wise Success Rates}

The instruction columns in Table~\ref{tab:robot-tabletop-results} specify the red or natural-wood base (Red/Wood), triangular or rounded roof (Triangular/Rounded), and same-color or cross-color pairing (Same/Cross). All routes share task-specific demonstrations, preprocessing, normalization, and adaptation budgets as described above.

For each task and route, let $s_{i,A},s_{i,B}\in\{0,1\}$ be the episode success indicators for the two instructions on layout $i$. We report
\begin{equation}
\begin{aligned}
    \mathrm{SR}_{c} &= \frac{100}{10}\sum_{i=1}^{10}s_{i,c},
    \qquad c\in\{A,B\}, \\
    \mathrm{Avg.\ SR} &= \frac{\mathrm{SR}_{A}+\mathrm{SR}_{B}}{2}.
\end{aligned}
\end{equation}
Each condition is evaluated on 10 layouts, so Avg.\ SR is also the percentage of successful episodes among all 20 trials for that task and route. The two condition scores reveal whether performance is concentrated on one instruction. Comparing instructions after independently resetting the same layout tests whether behavior follows the requested target rather than a fixed object or position preference. These success rates jointly reflect instruction following and execution ability. Across three task-specific checkpoints, each route has 60 evaluation episodes.
% TODO: Specify uncertainty estimation before reporting measured performance.

An episode succeeds when the instruction is fulfilled and the released roof remains supported without falling. For matching, both houses must be complete and both roof--body relations correct, with either order allowed. Unlike visual logical anomaly detection~\cite{peng2025samlad,peng2026vllmlad}, correctness here requires physically realizing the relation specified by the instruction. One completed house earns no episode-success credit; completion count (0, 1, or 2) is recorded separately. Ordinary task failures remain in the denominator of the corresponding instruction condition.

\end{document}